\documentclass{article}

\usepackage{microtype}
\usepackage{graphicx}
\usepackage{makecell}
\usepackage{subfigure}
\usepackage{booktabs} %
\usepackage{amssymb}%
\usepackage{pifont}%

\usepackage{hyperref}

\usepackage[accepted]{icml2020}
\usepackage{overpic}

\icmltitlerunning{
Normalization: Empirical Practices to Reap Some Benefits of Batch Norm without it}

\usepackage[utf8]{inputenc} %
\usepackage[T1]{fontenc}    %
\usepackage{hyperref}       %
\usepackage{url}            %
\usepackage{booktabs}       %
\usepackage{amsfonts}       %
\usepackage{nicefrac}       %
\usepackage{microtype}      %
\usepackage{amsmath}        %
\usepackage{mathtools}      %
\usepackage{array}          %
\usepackage{tablefootnote}  %
\usepackage{caption}
\usepackage{listings,multicol}       %

\usepackage{bbm}

\newcommand{\pp}[1]{\left( #1 \right)}

\newcolumntype{C}[1]{>{\centering\arraybackslash}m{#1}}
\usepackage{xcolor}
\definecolor{darkgreen}{rgb}{0,0.6,0}

\definecolor{darkblue}{rgb}{0,0,0.6}

\begin{document}

\twocolumn[
\icmltitle{Is Batch Norm unique? An empirical investigation and prescription to emulate the best properties of common normalizers without batch dependence%
}
\begin{icmlauthorlist}
\icmlauthor{Vinay Rao}{gr}
\icmlauthor{Jascha Sohl-Dickstein}{gr}
\end{icmlauthorlist}

\icmlaffiliation{gr}{Google Research, Mountain View, CA, USA}

\icmlcorrespondingauthor{Vinay Rao}{vinaysrao@google.com}
\vskip 0.3in
]

\icmlsetsymbol{equal}{*}
\printAffiliationsAndNotice{} 
\begin{abstract}
We perform an extensive empirical study of the statistical properties of Batch Norm and other common normalizers. 
This includes an examination of the correlation between representations of minibatches, gradient norms, and Hessian spectra both at initialization and over the course of training.
Through this analysis, we identify several statistical properties which appear linked to Batch Norm's superior performance. 
We propose two simple normalizers -- PreLayerNorm, and RegNorm -- which better match these desirable properties without involving operations along the batch dimension. 
We show that PreLayerNorm and RegNorm achieve much of the performance of Batch Norm without requiring batch dependence, that they reliably outperform LayerNorm, and that they can be applied in situations where Batch Norm is ineffective. 

\end{abstract}

\section{Introduction} \label{sec:intro}

Normalization has a long history in machine learning, and is central to the performance of Deep Neural Networks (DNNs). Traditional whitening transformations \citep{Kessy2018OptimalWA, Krizhevsky2009LearningML} and modern normalizers like Layer Norm \citep{LayerNorm}, Batch Norm \citep{BatchNorm} and others \citep{InstanceNorm,GroupNorm,shen2020rethinking,Salimans2016WeightNA} have helped train increasingly complex and deep DNN architectures \citep{He2016DeepRLResnet,Vaswani2017AttentionIA}. 
Batch Norm has been particularly effective \citep{Xiao2018DynamicalIA}.
Despite model performance being highly sensitive to choice of normalizer, there has been relatively little work exploring what properties of normalizers lead to good generalization performance. 
Although \citet{BatchNorm} proposed that Batch Norm helps by reducing Internal Covariate Shift (ICS), \citet{Santurkar2018HowDB} reported that it does not help ICS, but instead smooths the loss landscape. \citet{Bjorck2018UnderstandingBN, Luo2019TowardsUR} both found that Batch Norm allows large learning rates in specific settings, while \citet{Kohler2018TowardsAT} identified situations where Batch Norm accelerates training. \citet{Balduzzi2017TheSG} found that Batch Norm in ResNets allows deep gradient signal propagation in contrast to networks without it. \citet{Ghorbani2019AnII} studied the Hessian spectra of networks with Batch Norm and found that it helped prevent the appearance of large isolated eigenvalues in the spectrum. \citet{Yang2019AMF} studied the signal propagation of input representations and gradients through networks with Batch Norm and found that it causes gradient explosion under certain settings, while also causing exponential decay of information in deep networks without residual connections.
A frequent theme is an absence of comparison against other normalizers, raising the question of whether these behaviors or advantages are unique to Batch Norm.

Our contributions can be summarized as follows:
\begin{itemize}
    \item We study the effect of many normalization techniques on information propagation 
    \cite{schoenholz2016deep,Poole2016ExponentialEI}
    (Section \ref{subsec:infoprop}), loss surface curvature (Section \ref{sec:hessian}), and early training dynamics (Section \ref{sec:early_dynamics}).
    \item With extensive empirical evidence, we study the effect of depth, learning rate, and other hyperparameters and the behavior of normalization techniques under these changing settings. We also disentangle the importance of the mean subtraction and standard-deviation operations across the batch-dimension in Batch Norm. 
    \item Directly motivated by observed differences, we propose two simple alternatives to Batch Norm and Layer Norm that can match or outperform their generalization ability while closely matching likely beneficial statistical properties of Batch Norm (Sections \ref{sec:our_approaches} and \ref{sec:empirical}).
    \item  We show that these reach similar levels of performance as Batch Norm on datasets like Cifar10 and ImageNet, and improve on Layer Norm \citep{LayerNorm} for Transformer networks \citep{Vaswani2017AttentionIA} and recurrent architectures without resorting to additional complexity like Power Norm \citep{shen2020rethinking}.
    \item Finally, we share a set of failed approaches (Section \ref{sec:failed_exps}) we tried which did not improve model performance.
\end{itemize}

\section{Normalization techniques}

Let a deep neural network with $L$ layers have inputs $x^0$ of size (N,H,W,C) or (batch size, height, width, channels). Each layer performs an operation of the form
\begin{align}
x^l &= \phi\pp{ z^l }, \qquad z^l = W^l \circ x^{l-1} + b^l,
\end{align}
where $x^{l}$ is the output at layer $l$, $W^l$ and $b^l$ are layer $l$'s weight and bias parameters, $\circ \in$ \{Dense, Convolution\} indicates matrix multiplication or convolution, and $\phi$ is a nonlinear activation function like ReLU or tanh applied elementwise.
The normalization operations discussed below modify this layer equation to be
\begin{align}
    x^l &= \phi\pp{\gamma \tilde{z}^l + \beta}, \qquad \tilde{z}^l = \dfrac{z^l - \mu(\cdot)}{\sigma(\cdot)}
\end{align}
where the offset $\mu(\cdot)$ and scale $\sigma(\cdot)$ are computed in a different way for each normalizer. $\gamma$ and $\beta$ are 
post-normalization
scaling and 
 bias
parameters.

\textbf{Batch Norm} \citep{BatchNorm} requires additional overhead and complexity while training
to accumulate statistics for test-time, 
to synchronize these statistics across accelerators while training,
and to accumulate statistics over the batch dimension (involving non-local memory-access patterns). \citet{Ghorbani2019AnII} showed that using population mean and variance while training causes the Hessian to be ill-conditioned. We will show that this difference in test and train time behavior is crucial for the performance benefits of Batch Norm (Section \ref{sec:empirical}, Figure \ref{fig:batchsize}). 
For models with sequential inputs and outputs, the correct way to handle different length sequences within a batch, and more generally to handle the temporal axis, is not clear.

To overcome the difficulty of applying Batch Norm to sequential inputs/outputs, \citet{LayerNorm} introduced \textbf{Layer Norm},
which removes the dependency on the batch-size and has been used in achieving state of the art results with Transformers \cite{Vaswani2017AttentionIA, Ho2019AxialAITransformer1D}.
Weight Norm \cite{Salimans2016WeightNA} also removes the interaction between elements in the batch, while aiding good conditioning for optimization.

To disentangle the effects of subtracting the batch mean and the division by the batch standard deviation in our experiments, we consider two variants that ensure consistent scaling: Batch-Mean-Layer-Variance (\textbf{BMLV}) and Layer-Mean-Batch-Variance (\textbf{LMBV}). Table \ref{tab:norm_ops} summarizes these techniques in equation form.

Many other approaches have been proposed as alternatives to Batch Norm, including Group Norm \citep{GroupNorm}, Instance Norm \citep{InstanceNorm}, Power Norm \citep{shen2020rethinking} that improves performance of Transformer \citep{Vaswani2017AttentionIA} networks, Batch Renormalization \citep{Ioffe2017BatchRT} that reduces dependency on the minibatch,  and Norm Propagation \citep{Arpit2016NormalizationPA, Laurent2017RecurrentNP}.

\begin{table*}[h]
    \centering
    \begin{tabular}{c|c|c|c|c}
        Normalization & Layer Op & Norm Op & Mean Axes & Std.Dev Axes  \\
        \toprule
        Batch Norm & $x^l = \phi\pp{\gamma \tilde{z}^l + \beta}$ & $\tilde{z}^l = \dfrac{z^l - \mu^B}{\sigma^B}$ & $\mu^B: (N,H,W)$ & $\sigma^B: (N,H,W)$\\ \hline
        Layer Norm & $x^l = \phi\pp{\gamma \tilde{z}^l + \beta}$ & $\tilde{z}^l = \dfrac{z^l - \mu^i}{\sigma^i}$ & $\mu^i: (H,W,C)$ & $\sigma^i: (H,W,C)$\\ \hline
        Weight Norm & $x^l = \dfrac{W^l \circ x^{l-1}}{||W^l||_F} + b^l$ & - & - & - \\ \hline
        \midrule
        (Ours)\\
        BMLV & $x^l = \phi\pp{\gamma \tilde{z}^l + \beta}$& $\tilde{z}^l = \dfrac{z^l - \mu^B}{\sigma^i}$ & $\mu^B: (N,H,W)$ & $\sigma^i: (H,W,C)$\\ \hline
        LMBV & $x^l = \phi\pp{\gamma \tilde{z}^l + \beta}$ & $\tilde{z}^l = \dfrac{z^l - \mu^i}{\sigma^B}$ & $\mu^i: (H,W,C)$ & $\sigma^B: (N,H,W)$\\ \hline
        \midrule
        (Ours, improved)\\
        PreLayer Norm & $x^l = \phi\pp{\gamma \hat{z}^l + \beta}$ & \makecell{$\hat{z}^l = \dfrac{\bar{z}^l}{\sigma^i(\cdot)}$, \\ $\bar{z}^l = W^l \circ \pp{x^{l-1}-\mu^i(x^{l-1})}$} & $\mu^i:(H,W,C)$ & $\sigma^i:(H,W,C)$\\ \hline
        RegNorm* & $x^l = \phi\pp{ \gamma \bar{z}^l + \beta}$ & \makecell{$\bar{z}^l = \dfrac{z^l}{\sigma^i(\cdot)}, z^l = W^l \circ x^{l-1}$} & - & $\sigma^i:(H,W,C)$\\
        \bottomrule
    \end{tabular}
    \caption[]{Specifications for the the normalization schemes we consider in this paper. *RegNorm additionally uses a regularization term $r(\Bar{z}) = \mathbb{E}_{a,b}\left[ \sum_i((\Bar{z}^a_i + \Bar{z}^b_i)^2 - 2) \right]$ where $a, b \in Batch$ and $i \in (H, W, C)$ (Section \ref{sec:reg_norm}).}
    \label{tab:norm_ops}
\end{table*}

\section{Improved normalizers}  \label{sec:our_approaches}

We propose two new normalizers which better match the statistical properties and good performance of Batch Norm without requiring operations over the batch dimension or differences in test and train time behavior. 
See Table \ref{tab:norm_ops} for additional specification. Their design will be further motivated by the experimental analysis of existing normalizers presented below. 
Qualitative properties of different normalizers are compared in Table \ref{tab:compare_against_bn}.

\subsection{Pre-Normalization:PreLayerNorm}
A simple shift in the order of normalization operations that we term Pre-Normalization, helps mimic some properties of Batch Norm without batch-axis operations.

Particularly, we study PreLayerNorm, where the normalizer subtracts the layer's mean $\mu^i$ before the Affine operation and divides the layer's standard-deviation $\sigma^i$ before the nonlinearity $\phi$.

To see why PreLayerNorm can be expected to behave differently than Layer Norm, consider the affine transformation $z^l = W^l \circ x^{l-1}$ where $W^l \in R^{N_{l-1} \times N_l}$ is initialized from an isotropic Gaussian. 
Even if $x^{l-1}$ has non-zero mean across units, in expectation $z^l$ will be zero-mean across units, due to the randomness in the weight matrix $W^l$. Therefore, subtracting the mean across units can be expected to have a far smaller effect if applied after multiplication by $W^l$. 
To the degree that the mean across units in $x^{l-1}$ resembles the mean across the batch, Pre-Normalization more closely emulates the behavior of Batch Norm.

\subsection{Regularized Normalization (RegNorm)} \label{sec:reg_norm}
We propose a normalization scheme that in conjunction with a regularization term added to the loss, ensures that the mean of the activations across the batch is 0 while ensuring consistent scaling. At initialization, this technique mimics the information propagation properties of Layer Norm, while after minimizing the regularizer it mimics those of BMLV.
The regularization term is
\begin{align}
    r(\Bar{z}) &= \mathbb{E}_{a,b}\left[ \sum_i((\Bar{z}^a_i + \Bar{z}^b_i)^2 - 2) \right]
    ,
\end{align}
where $a, b \in Batch$ and $i \in (H, W, C)$. Note that the regularizer is computed after the inputs are normalized, and so will not depend on the activation norm. 
The loss-function used during model training is modified to be $L' = L + \lambda \sum_{l}r^l$, where the sum is over network layers, and $\lambda$ is a regularization strength hyperparameter. This regularizer has a unique minimum when the mean of the activations across the batch is 0 -- see Appendix \ref{app:regularizer_proof} for a proof. We further support this empirically in Figure \ref{fig:reg_coeff}.

\section{Properties at initialization}\label{sec:init}
In this section we analyze the statistical properties of networks at initialization, under different normalization techniques. \citet{Xiao2019DisentanglingTA} studied the effect of information propagation on trainability and generalization, showing that while  networks in the chaotic phase (initially similar samples become dissimilar with depth)
train faster, their generalization suffers. Conversely, networks in the ordered phase (the correlation between samples converges to 1 with depth) are harder to train, but show better generalization. Normalization techniques induce different information propagation behavior. We first examine this behavior at initialization. In Section \ref{sec:early_dynamics}, we will see how these phases evolve quickly through the early stages of training.

\subsection{Information propagation through layers} \label{subsec:infoprop}
Information propagation \citep{schoenholz2016deep,Poole2016ExponentialEI} is the measure of how similar the representations of two inputs are through a DNN. We calculate this as the correlation between two similar minibatches -- $\operatorname{corr}(x+\epsilon_1, x+\epsilon_2)$, where $\epsilon_1,\epsilon_2 \sim \mathcal{N}(0,\sigma)$, and $x$ here represents an entire minibatch of data. If the correlation approaches 1 after the application of many layers, then dissimilar minibatches are converging on the same activation patterns. On the other hand if the correlation converges towards 0 (or other value less than 1), then the network is chaotic or behaves like a hash map, where inputs with initially minor differences are mapped to very different points in the representation space.

\begin{figure*}[h]
\centering
  \begin{overpic}[width=0.4\linewidth]{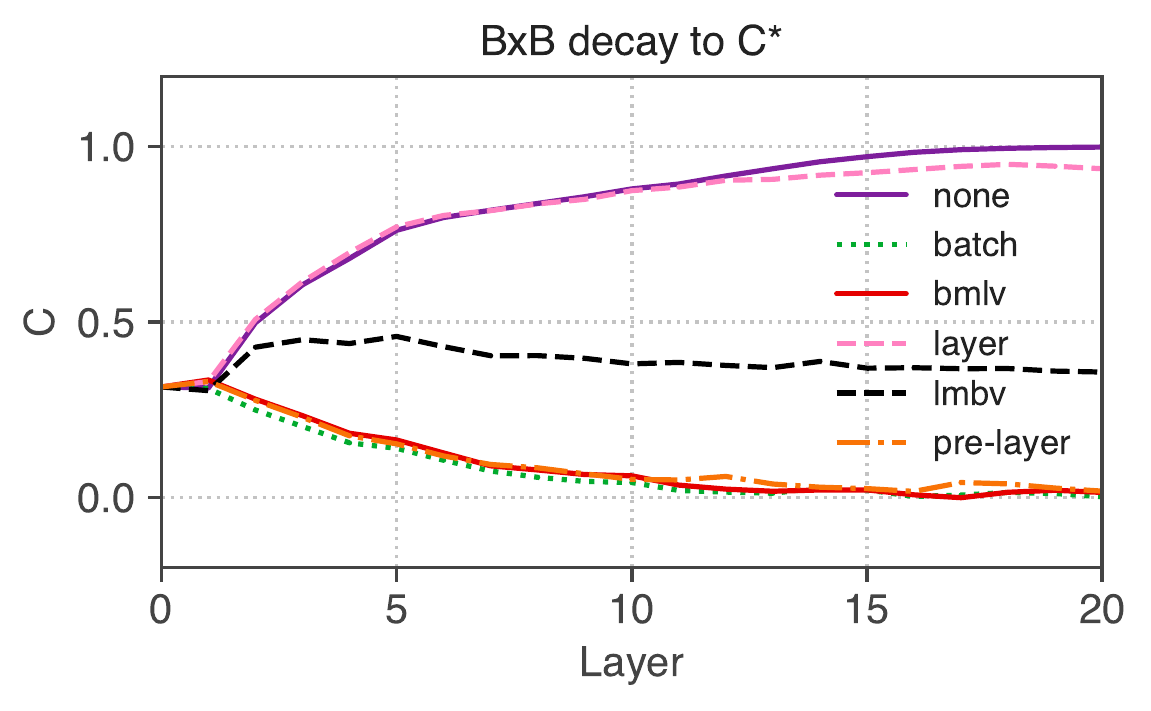}
 \put (2,3) {\textbf{\small(a)}}
 \label{fig:fc_bxb}
\end{overpic}    
\qquad
  \begin{overpic}[width=0.4\linewidth]{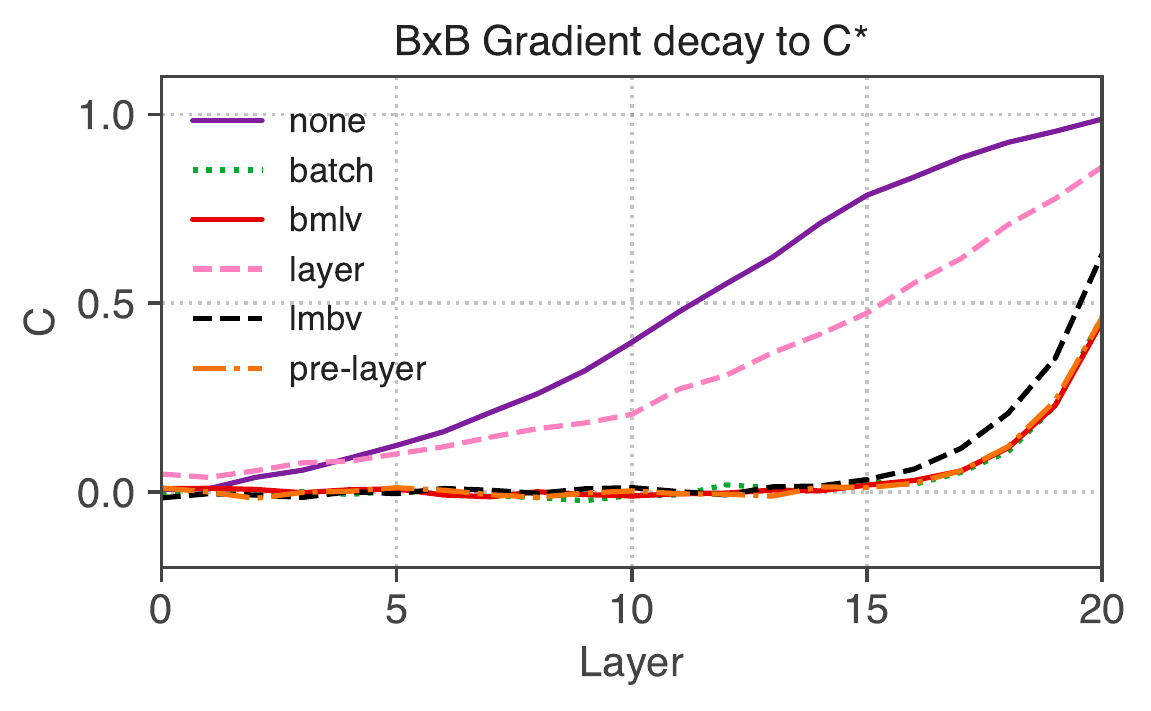}
 \put (2,3) {\textbf{\small(b)}}
 \label{fig:grad_fc_bxb}
\end{overpic}    
\caption{\small{Correlation of input representations {\em (a)} and gradients {\em (b)} between two similar minibatches through 20 layers of a fully-connected DNN with ReLU activation. {\em (a)} Batch Norm decorrelates the representations through depth (as do BMLV, PreLayerNorm) while LMBV preserves the correlation. The representations become highly correlated for Layer Norm and No Norm. {\em (b)} While all normalizers cause gradients to become decorrelated at the first layer, Batch Norm, BMLV and PreLayerNorm cause faster decorrelation.}
}
\label{fig:fc_bxb_grad}
\end{figure*}

\begin{figure*}[h]
    \centering
    \begin{overpic}[width=0.4\linewidth]{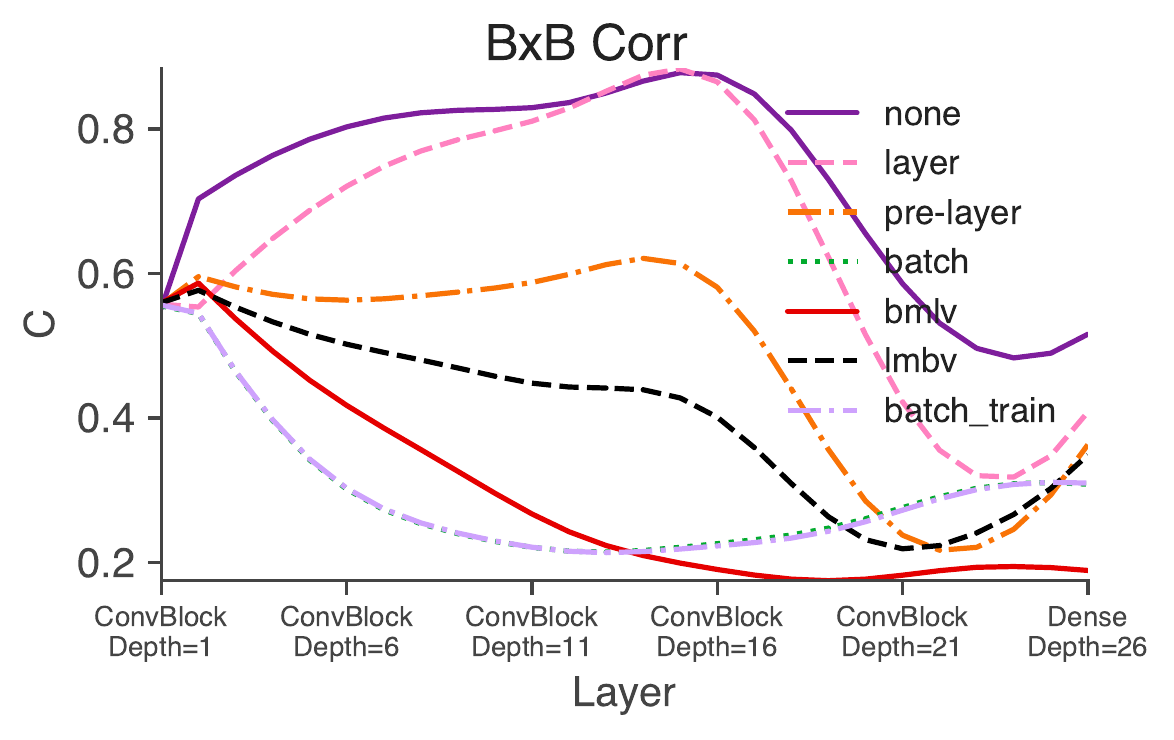}
    \put (2,3) {\textbf{\small(a)}}
    \label{fig:wrn_bxb_noskip}
    \end{overpic}
    \qquad
    \begin{overpic}[width=0.4\linewidth]{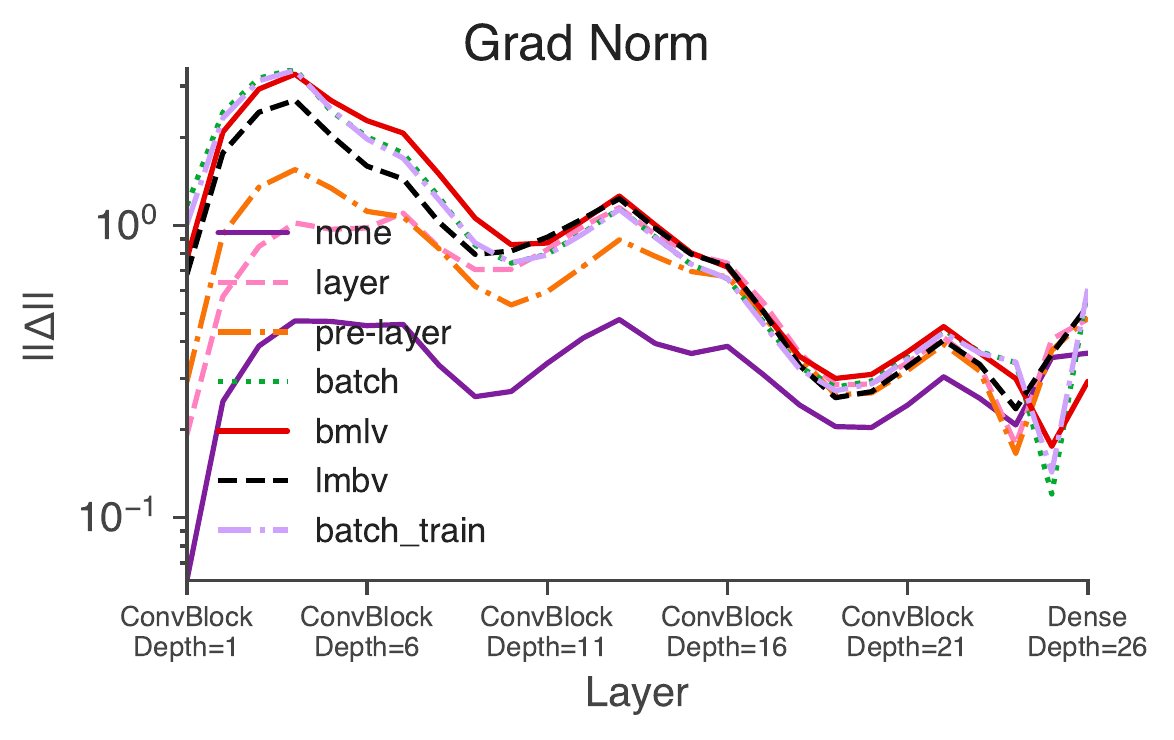}
    \put (2,3) {\textbf{\small(b)}}
    \label{fig:grad_norm_noskip}
    \end{overpic}
 \caption{\small{Correlation of inputs {\em (a)} and L2-Norm of gradients {\em (b)} between similar minibatches through a 26-layer WideResnet with ReLU activations and no skip connections.
 {\em (a)} Batch Norm (and BMLV) decorrelates the representations through depth while Layer Norm leads to highly correlated representations. Using PreLayerNorm leads to more decorrelation and greater similarity to Batch Norm.
 {\em (b)} Batch Norm (and BMLV) induces the most gradient explosion at initialization, while No Norm induces the least. PreLayerNorm induces more gradient explosion than Layer Norm.}
 }
    \label{fig:wrn_bxb_grad_norm}
\end{figure*}

We first plot the propagation of MNIST inputs through a fully-connected network with ReLU activations in Figure \ref{fig:fc_bxb_grad}(a). In Figure \ref{fig:wrn_bxb_grad_norm}(a) we consider a WideResnet~\citep{Zagoruyko2016WideRN} with 26 layers and no skip connections, with a channel multiplier set to 2, and propagate batches from the Cifar-10 dataset \citep{cifar10}. The majority of experiments in the paper will use this WideResnet architecture, with experiments including skip connections in the Appendix.

We find that Batch Norm preserves less information about the input than any other standard normalization technique. 
Skip connections improve information propagation (Figure \ref{fig:wrn_bxb}), but don't change the relative behavior of Batch Norm compared with other normalizers. 
The BMLV and LMBV conditions suggest that for a ReLU network the subtraction of the mean across the batch is the primary driver of correlation decay, and division by the standard deviation across the batch plays little role.

\subsection{Correlation of gradients through layers}
We examine the correlation of gradients through the same fully-connected network in Figure \ref{fig:fc_bxb_grad}(b). \citet{Sankararaman2019GradConfusion} postulate that gradient confusion, or highly-correlated gradients causes slower training. We notice that Batch Norm produces the most decorrelated gradients, while Layer Norm produces highly-correlated gradients.

\subsection{Gradient explosion through layers}
\citet{Yang2019AMF} predicted that Batch Norm causes gradient explosion
in DNNs, when repeatedly applied in series.
We empirically observe in Figure \ref{fig:grad_norm} that with skip connections, this phenomenon does not significantly occur (note that this \emph{is} consistent with the theory result). 
However, without skip connections (Figure \ref{fig:wrn_bxb_grad_norm}(b)), we observe that Batch Norm and closely related techniques do cause more growth in gradients with depth compared to other techniques.

\section{Early training dynamics} \label{sec:early_dynamics}
\citet{lewkowycz2020large,Frankle2020EarlyPhase, Jastrzebski2020TheBP} observed that DNNs undergo important changes in the early stages of training. 
We examine the behavior of the properties from Section \ref{sec:init} over the course of training, to understand how the effect of different normalizers evolves over the course of training.

Using Batch Norm as a yardstick, we see how our proposed techniques compare to its early training dynamics. Unless explicitly stated, we plot the dynamics for DNNs without skip/residual connections (see Appendix \ref{appendix:ablation} for experiments with residual connections). 
We perform a grid search $[10^{-3},10^{-2},\dots,10]$ over learning rates for each of the approaches while generating these plots. 
We also discuss the surprising transition of Batch Normalization away from inducing chaos at initialization -- underscoring the importance of the first few steps of training.

\begin{figure*}[]

\centering
    \begin{overpic}[width=0.49\linewidth]{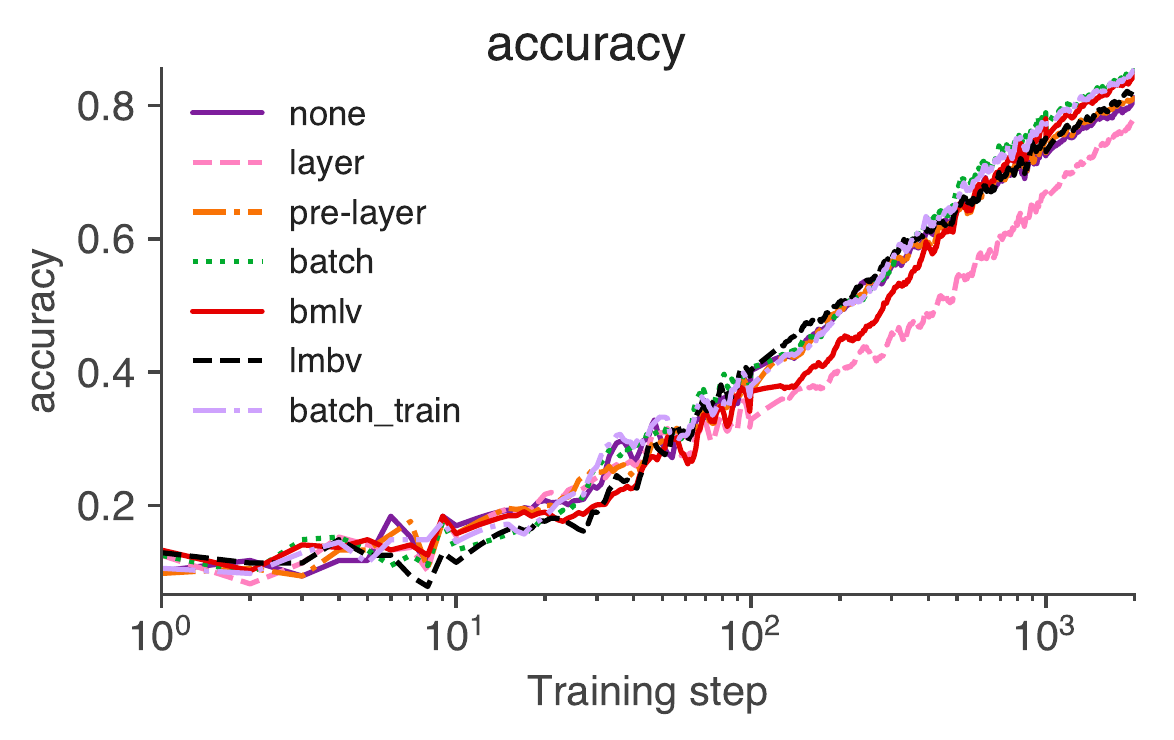}
    \put (6,4) {\textbf{\small(a)}}
    \label{fig:train_accuracy_noskip}
    \end{overpic}
    \begin{overpic}[width=0.49\linewidth]{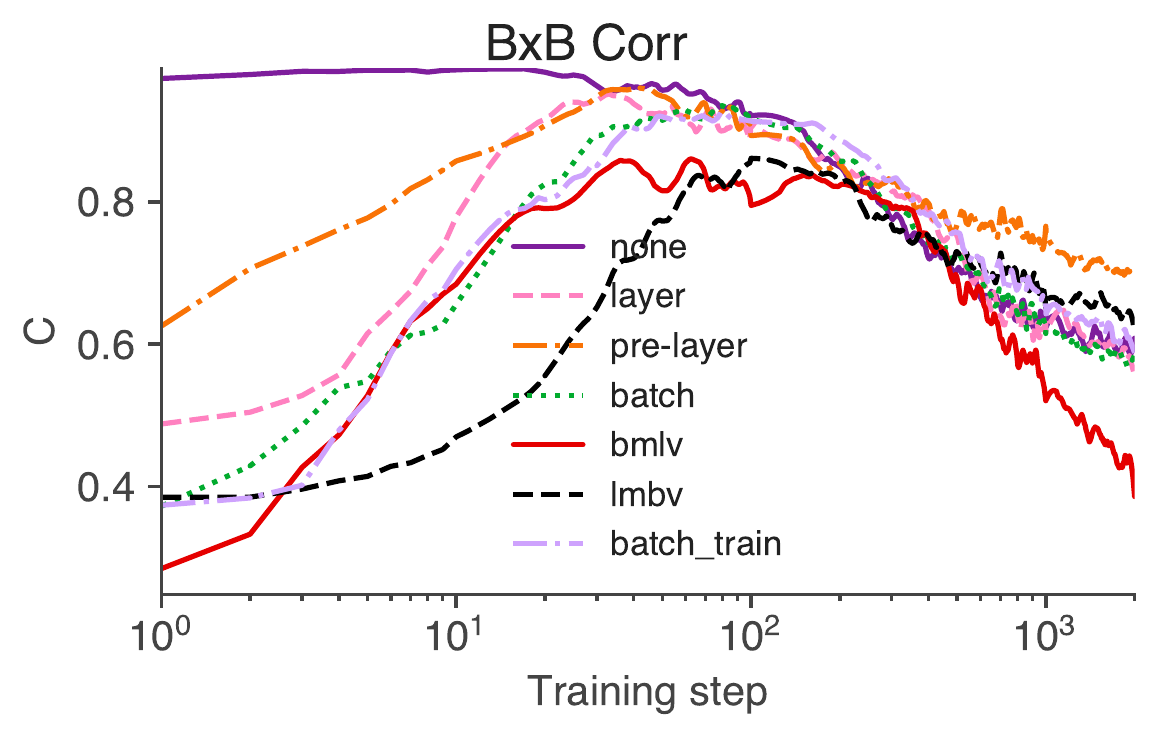}
    \label{fig:train_bxb_corr}
    \put (6,4) {\textbf{\small(b)}}
    \end{overpic}
\quad
    \begin{overpic}[width=0.49\linewidth]{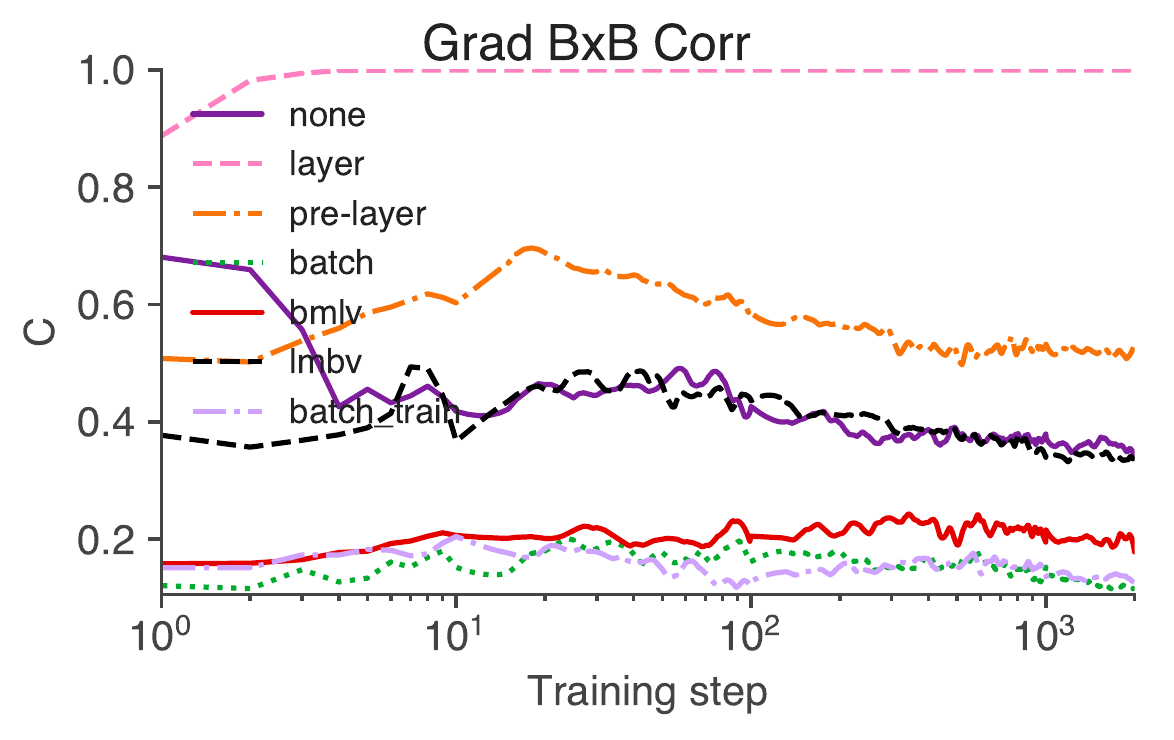}
    \label{fig:train_bxb_corr_grad}
    \put (6,4) {\textbf{\small(c)}}
\end{overpic}
    \begin{overpic}[width=0.49\linewidth]{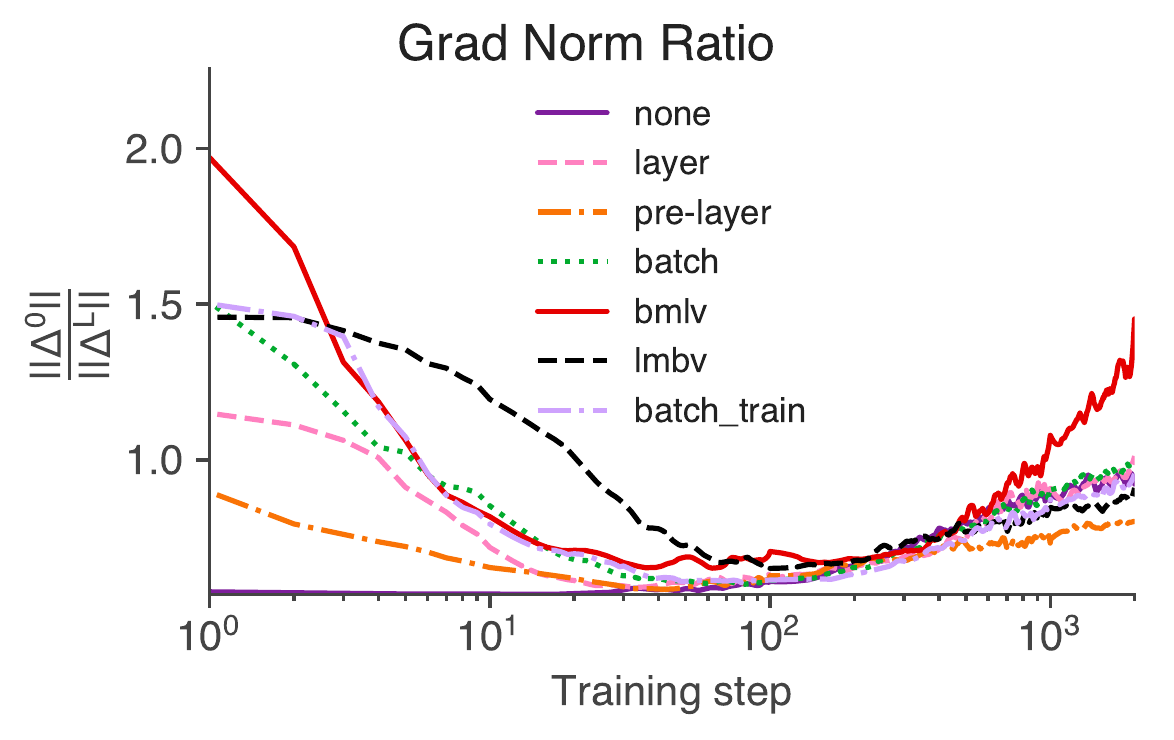}
    \label{fig:train_grad_explode_noskip}
    \put (6,4) {\textbf{\small(d)}}
\end{overpic}

 \caption{{\em (a)} Accuracy, {\em (b)} correlation of output representations, {\em (c)} correlation of input gradients, and {\em (d)} ratio of L2-Norm of the gradients at the first layer to the last layer of a 26-layer WideResnet with ReLU activations and no skip connections through the first 2000 steps of training on Cifar10, with each technique trained using tuned learning rates. {\em (b)} In networks with any normalization, DNNs undergo a transition from being more chaotic (with Batch Norm and BMLV the most chaotic) with decorrelated representations, to more ordered with correlated representations. The correlation between representations then gradually decreases out to 2000 training steps. Note that No Norm only undergoes one transition from highly correlated to decorrelated. {\em (c)} Batch Norm (and BMLV) has the least correlated gradients or gradient confusion, while Layer Norm has the most highly correlated gradients. PreLayerNorm in this case does not fully match Batch Norm's behavior. {\em (d)} Batch Norm and BMLV induce the most gradient explosion in the first phase of early training while Layer and PreLayerNorm cause lower gradient explosion. No Norm causes the gradients to vanish. All the techniques undergo a shift where there is an increasing amount of gradient explosion later in the first 2000 training steps.
 \label{fig:training_dynamics}
 }
    
\end{figure*}

From Figure \ref{fig:training_dynamics}(a), we observe that \textbf{ accuracy trajectories} look similar for most normalizers, with layer norm lagging other approaches. 
By step 1000, networks with all normalization schemes perform with greater than 50\% accuracy.

We plot the \textbf{correlations of the output}s at the final pre-activation layer between similar batches as described in Section \ref{subsec:infoprop} through training (Figure \ref{fig:training_dynamics}(b)). Even though Batch Norm is chaotic at initialization where all inputs are mapped to very different points in representation space, there is a phase where the representations become more correlated, before gradually decorrelating again. 
Additionally, Layer Norm causes its input representations to be highly correlated through training. This might prevent it from exploring representation spaces. Using our proposed Pre-Normalization technique, we are able to get closer to Batch Norm's behavior without any operations across the batch-dimension.

\citet{Sankararaman2019GradConfusion} proposed the concept of \textbf{Gradient Confusion}, where having highly similar gradients between batches leads to slower training with SGD. We note from Figure \ref{fig:training_dynamics}(c) that Batch Norm induces the least gradient confusion. However, BMLV behaves similarly. Further, while Layer Norm has \textbf{gradient correlations} that rapidly approach 1, PreLayerNorm maintains intermediate values of gradient correlation.

\begin{figure*}[]
    \centering
    \includegraphics[width=\textwidth]{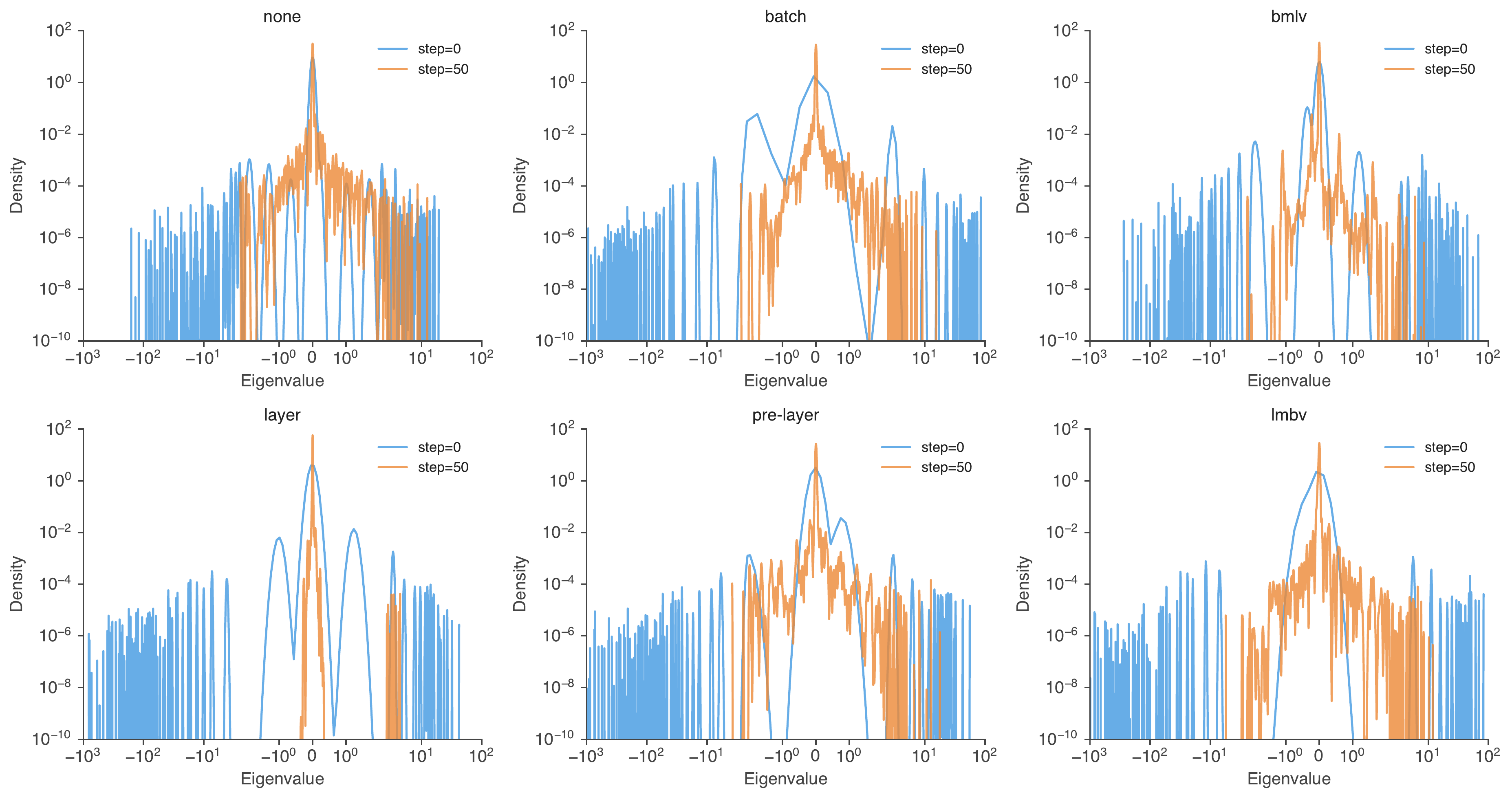}
    \caption{Density of Eigenvalues of the Hessian (cross-entropy) at initialization and 50 steps into training (we used the same learning rate for all techniques) for a 26-layer WideResnet with ReLU activations and no skip connections. Batch Norm has a wider spectrum at initialization. After 50 steps, Layer Norm's spectrum has shrunk and contains outliers. PreLayerNorm behaves similarly to Batch Norm.}
    \label{fig:density_noskip}
\end{figure*}
\section{Hessian spectrum analysis} \label{sec:hessian}
In this section, we highlight some properties of the Hessian spectrum of the WideResnet without skip connections described before. We defer the DNN with skip connections to the appendix (Figure \ref{fig:density_skip}) because it shows qualitatively similar relationships between normalizers.
We estimate the Hessian for the cross-entropy loss using the Lanczos algorithm as described in \citet{Ghorbani2019AnII} using 512 samples of Cifar-10.

We visualize the \textbf{density of eigenvalues} of the Hessian 50 steps into training in Figure \ref{fig:density_noskip}. 
We observed that very quickly through training, the spectrum undergoes a transformation (and we note that this is predictive of its shape very late in training), where Batch Norm suppresses its outliers and does not have large negative eigenvalues. Layer Norm has a small spectrum, with many outliers which has been noted \cite{Ghorbani2019AnII} to cause slow training. Our proposal PreLayerNorm is able to condition the Hessian similarly to Batch Norm.
\begin{figure*}[]
    \centering
    \includegraphics[width=\textwidth]{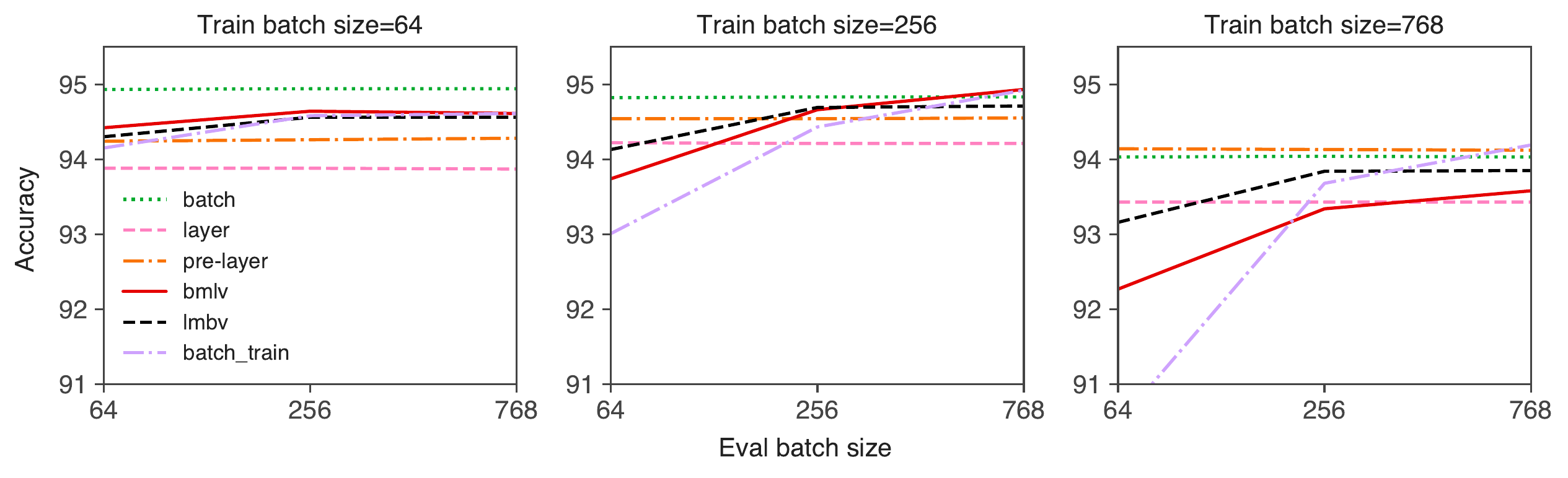}
    \caption{Performance of networks trained with different batch sizes, across a variety of test-time batch sizes. For LMBV, BMLV and Batch-Train, we did not use accumulated statistics. We kept the learning rate constant across batch-sizes for this experiment. Without accumulated statistics (i.e Batch-Train), Batch Norm's performance is highly dependent on batch size. PreLayerNorm outperforms Layer Norm in all the cases.}
    \label{fig:batchsize}
\end{figure*}

\begin{figure*}[]
    \centering
    \begin{overpic}[width=0.49\linewidth]{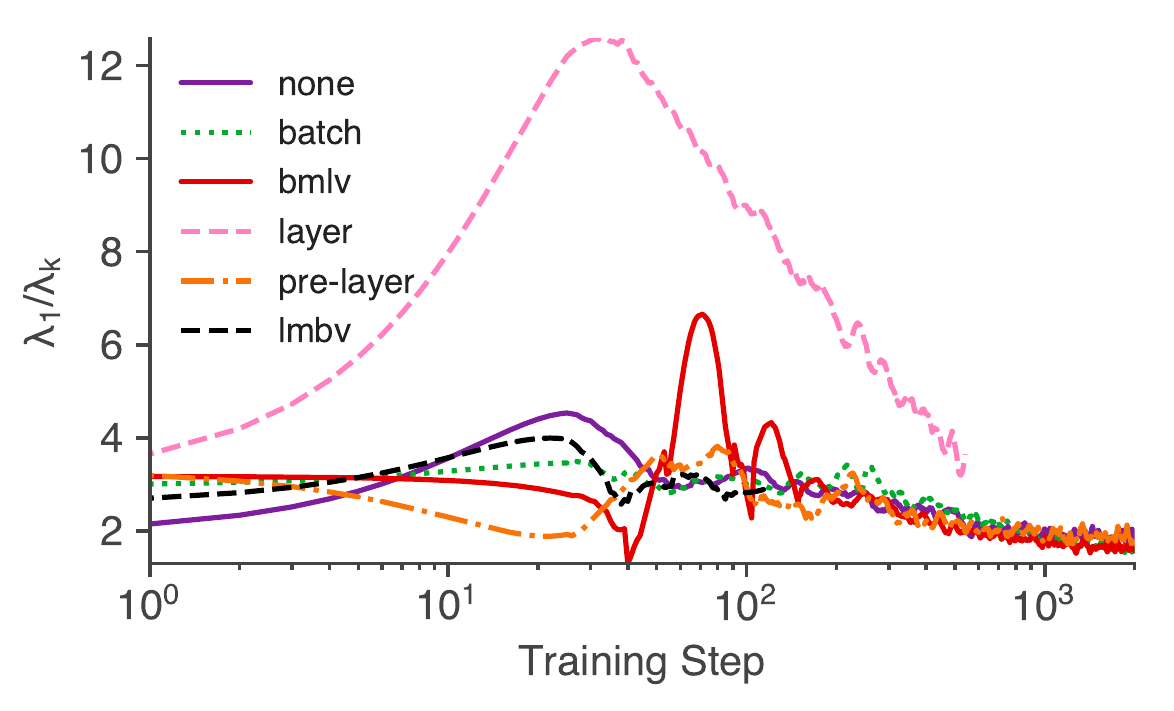}
    \label{fig:lambda1_lambdak_noskip}
    \put (6,4) {\textbf{\small(a)}}
    \end{overpic}
    \begin{overpic}[width=0.49\linewidth]{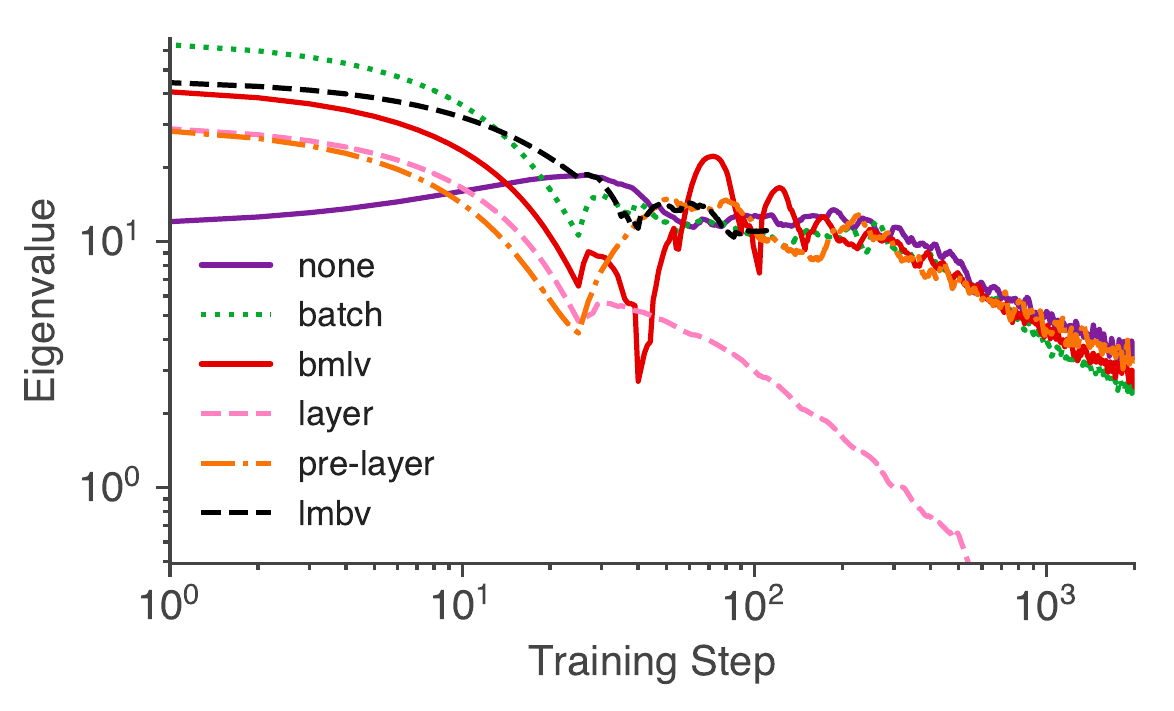}
    \label{fig:lambdamax_noskip}
    \put (6,4) {\textbf{\small(b)}}
    \end{overpic}
    \caption{\small{Ratio of the largest eigenvalue of the Hessian (cross-entropy) to the K-th Eigenvalue {\em (a)} and the largest eigenvalue {\em (b)} through training a 26-layer WideResnet with ReLU activations and no skip connections. We set K=10 to correspond to the 10 output classes used in Cifar10. {\em (a)} All techniques including no normalization are able to suppress outlying eigenvalues, except for Layer Norm. {\em (b)} Although the different techniques differ at initialization, they seem to be scaled and conditioned similarly later in training. 
    }
    \label{fig:eigenvalue_stuff}
    }
\end{figure*}

\citet{Santurkar2018HowDB} proposed that Batch Norm helps optimization by making the loss-surface smoother. Further, according to \citet{Ghorbani2019AnII} Batch Norm helps by suppressing outliers in the spectrum. This is measured by the \textbf{ratio} of the largest eigenvalue ($\lambda_1$) to the \textit{K}-th eigenvalue ($\lambda_k$), \boldmath{$\lambda_1 / \lambda_k$}\unboldmath{}. We set \textit{K}=10 to correspond to the number of output classes in Cifar-10, but it should be noted that this is a heuristic guess. We notice from Figure \ref{fig:eigenvalue_stuff}(a) that in fact most normalization techniques are able to suppress outliers, except Layer Norm. This might explain the discrepancy in training convolutional architectures with Layer Norm. However, we are able to overcome this with PreLayerNorm without any operations across the batch-axis.

The largest stable learning rate is roughly inversely proportional to the \textbf{largest eigenvalue} of the Hessian \citep{LeCun1992AutomaticLR} (see \citet{lewkowycz2020large} for subtleties). 
At initialization, it seems from Figure \ref{fig:eigenvalue_stuff}(b) that Batch Norm must allow the lowest learning rate, however in conjunction with suppressing outliers as seen previously, the maximum eigenvalue quickly decreases, allowing the possibility of larger learning rates. We further note that most other regularizers condition the spectrum similarly through training. LayerNorm is unable to suppress outliers and has a small spectrum that does not allow one to take advantage of its low eigenvalues to use larger learning rates.

\begin{table*}[h]
\begin{tabular}{ccccccc}
     \toprule
     Desired Property & Batch Norm & BMLV & Layer & PreLayerNorm & RegNorm \\
     \midrule
     Consistent scaling and centering & + & + & + & + & + \\
     Faster training and better generalization & + & + & -* & + & + \\
     Allow training deeper networks & + & + & + & + & + \\
     Outlier eigenvalue suppression through training & + & + & - & + & + \\
     No dependence on batch size & - & - & + & + & + \\
     Consistent training and test behavior & - & - & + & + & + \\
     Easy application to sequential or recurrent models & - & - & + & + & + \\
     Reduce gradient confusion through training & + & + & - & $\sim$ & $\sim$ \\
     Transition from chaotic to ordered during early training & + & + & + & + & + \\
     Induces gradient explosion$^\dagger$ & + & + & - & - & - \\
     \bottomrule
\end{tabular}
\caption{Several desirable properties for normalizers on DNNs are listed in this table, alongside the methods compared in this paper. A `+' indicates that the technique induces the listed effect, while `-' indicates that it does not.\\
\small $^*$ Layer Norm enables better generalization in Transformer networks, but does not have the same effects in Deep Convolutional Networks in our experiments.
$^\dagger$ It remains unclear whether or not this is a desirable property.
$^\sim$ While PreLayerNorm and Regularized Norm are able to reduce gradient confusion, they do not do so to the extent that Batch Norm does.
}
\label{tab:compare_against_bn}
\end{table*}
\begin{table*}
    \centering
    \begin{tabular}{ccccc}
        \textbf{Norm Type} & \textbf{Cifar10} & \textbf{ImageNet} & \textbf{ImageNet} & \textbf{LM1B}\\
        & Wide-Resnet & Resnet50-v1 & Resnet101-v1 & Transformer-1D\\
         & (Accuracy) & (Accuracy) & (Accuracy) & (Perplexity)\\
        \toprule
        None & 0.0 & 71.254 & 73.153 & 59.22\\
        Batch & \textbf{95.953} & \textbf{76.293} & \textbf{78.104} & - \\
        Batch-Train & 95.512 & 74.780 & 76.693 & -\\
        Layer & 95.653 & 66.623 & 75.745 & 38.341\\
        PreLayerNorm & 95.733 & 75.107 & \underline{77.655} & \textbf{\underline{38.002}}\\
        RegNorm & \underline{95.843} & 74.782 & 76.725 & -\\
        PreRegNorm* & 95.813 & \underline{75.878} & 77.452 & -\\
        BMLV & 95.432 & 75.174 & 76.814 & -\\
        LMBV & 95.392 & 74.837 & 76.087 & -\\
        \bottomrule
    \end{tabular}
    \caption{Performance using different normalizers of a WideResnet(26,10) on Cifar10, Resnet50-v1 on ImageNet, Resnet101-v1 on ImageNet and a Transformer1D on LM1B.\\
    \small $^*$ PreRegNorm is defined as RegNorm with layer mean subtraction before the Affine operation.}
    \label{tab:cifar_imagenet_results}
\end{table*}

\section{Performance on realistic tasks} \label{sec:empirical}

We compare RegNorm and PreLayerNorm against Batch Norm and Layer Norm on several existing tasks. Additional experiments are discussed in Appendix \ref{appendix:ablation}.

In all experiments, we replace the default normalization technique used with our alternatives, and perform identical hyperparameter tuning.

Table \ref{tab:cifar_imagenet_results} shows the performance of a 26-layer WideResnet \cite{Zagoruyko2016WideRN} with channel-multiplier set to 10 on Cifar-10. The best performing technique, Batch Norm is highlighted in \textbf{bold} while the best amongst our proposed techniques is \underline{underlined}. Note that our techniques are able to perform better than Batch Norm if it does not make use of aggregated mean and variance during test-time (Batch-Train). Further, using Pre-Normalization helps increase performance.

We further compare the performance of these techniques using the Resnet-V1 \cite{He2016DeepRLResnet} architecture on ImageNet. Table \ref{tab:cifar_imagenet_results} shows the results with 50 and 101 layers. Although Batch Norm performs best, we close the gap to it without any additional computation and consistently outperform Batch-Train. Further we note that using Pre-Normalization helps increase performance.

Using a Transformer architecture proposed in \citet{Ho2019AxialAITransformer1D}, we compare PreLayerNorm against Layer Norm and no normalization. Note that in the case of sequential inputs or outputs, operations across the batch-axis are confounding because they introduce the need to use masking and limit the ways in which mean and variance can be aggregated for all the timesteps. There has been previous discussion of the placement of Layer Norm in these architectures in \citet{Xiong2020OnLN}. By simply replacing Layer Norm with Pre-Normalization, we are able to outperform it on the LM1B language modeling task (Table \ref{tab:cifar_imagenet_results}).

 From Figures \ref{fig:fc_bxb_grad}, \ref{fig:wrn_bxb_grad_norm} we see that BMLV most closely matches Batch Norm's statistical properties at initialization, and through Figures \ref{fig:training_dynamics}(a,b,c,d) we see again that BMLV most closely resembles Batch Norm's statistical and gradient dynamics through the early stages of training. For these networks with ReLU nonlinearities, we notice that the Batch Mean subtraction is more important, further motivating the design of RegNorm which emulates the behavior of BMLV in a trained network.
 
\section{Batch Norm and dependence on batch-size}
With increasing batch size, Batch Norm causes slower decorrelation of input representations through depth at initialization \citep{Yang2019AMF}. 
The stochastic choice of batch members induces noise in the Batch Norm estimation of the mean and variance.
This causes discrepancy in the training and test-time behavior. 
As we observe in Figure \ref{fig:batchsize}, the performance of Batch Norm and other methods that have operations across the batch change with the batch size if the population mean and variance are not aggregated for use at test-time. Even for large test-time batch sizes, Batch Norm without aggregated means and variances produces significantly worse test accuracy (compare $\operatorname{batch_train}$ to $\operatorname{batch}$).  The presence of batch noise during training, and accumulated statistics at test time, seems to be crucial to Batch Norm's better performance than other methods.

\section{Discussion}

We performed an empirical study of Batch Norm and competing normalizers, examining information propagation and Hessian spectra both at initialization and throughout training. 
From analyzing the loss surface curvature, we observe that all normalization schemes (except Layer Norm) are able to suppress outlying eigenvalues similarly. While Weight Norm does not affect the conditioning of the Hessian, introducing residual/skip connections improves conditioning similarly across all techniques.
Batch Norm induces poor conditioning while training if true aggregated mean and variance are used while training \cite{Ghorbani2019AnII}. However, as we empirically observed previously (Sec \ref{sec:empirical}), it also relies on these aggregated metrics to outperform other techniques while testing. This hints that it is relying on the regularization induced by the noise in the mean and variance estimates. Note in Section \ref{sec:failed_exps} that our attempts to artificially inject regularizing noise did not improve performance.
We disentangled the effect of the mean subtraction and standard-deviation division operations by studying BMLV and LMBV and showed that batch mean subtraction plays a bigger role than batch standard-deviation division.

We further introduce two new normalizers -- RegNorm and PreLayerNorm -- that  better match the statistical and conditioning properties of Batch Norm without relying on operations across the batch. These normalizers perform nearly as well as Batch Norm in cases where Batch Norm is commonly used.
We compare these techniques against Batch Norm in Table \ref{tab:compare_against_bn}. 
These new techniques further outperform Layer Norm for models such as Transformers, with sequential inputs/outputs, where Batch Norm is ineffective. Additionally, PreLayer Norm shows better Hessian conditioning in cases where using Layer Norm does not.

As a service to future researchers, we provide a summary of failed experiments in Appendix \ref{sec:failed_exps}.

\section*{Acknowledgments}

We thank
Greg Yang,
Justin Gilmer,
David Page,
Jeffrey Pennington,
and Sam Schoenholz
for useful discussion and feedback.

\clearpage
\newpage

\bibliographystyle{icml2020}
\bibliography{main}

\begin{thebibliography}{34}
\providecommand{\natexlab}[1]{#1}
\providecommand{\url}[1]{\texttt{#1}}
\expandafter\ifx\csname urlstyle\endcsname\relax
  \providecommand{\doi}[1]{doi: #1}\else
  \providecommand{\doi}{doi: \begingroup \urlstyle{rm}\Url}\fi

\bibitem[Arpit et~al.(2016)Arpit, Zhou, Kota, and
  Govindaraju]{Arpit2016NormalizationPA}
Arpit, D., Zhou, Y., Kota, B.~U., and Govindaraju, V.
\newblock Normalization propagation: A parametric technique for removing
  internal covariate shift in deep networks.
\newblock \emph{ArXiv}, abs/1603.01431, 2016.

\bibitem[Ba et~al.(2016)Ba, Kiros, and Hinton]{LayerNorm}
Ba, J., Kiros, J.~R., and Hinton, G.~E.
\newblock Layer normalization.
\newblock \emph{ArXiv}, abs/1607.06450, 2016.

\bibitem[Balduzzi et~al.(2017)Balduzzi, Frean, Leary, Lewis, Ma, and
  McWilliams]{Balduzzi2017TheSG}
Balduzzi, D., Frean, M., Leary, L., Lewis, J.~P., Ma, K. W.-D., and McWilliams,
  B.
\newblock The shattered gradients problem: If resnets are the answer, then what
  is the question?
\newblock \emph{ArXiv}, abs/1702.08591, 2017.

\bibitem[Bjorck et~al.(2018)Bjorck, Gomes, and
  Selman]{Bjorck2018UnderstandingBN}
Bjorck, J., Gomes, C.~P., and Selman, B.
\newblock Understanding batch normalization.
\newblock In \emph{NeurIPS}, 2018.

\bibitem[Frankle et~al.(2020)Frankle, Schwab, and
  Morcos]{Frankle2020EarlyPhase}
Frankle, J., Schwab, D.~J., and Morcos, A.~S.
\newblock The early phase of neural network training.
\newblock \emph{ArXiv}, abs/2002.10365, 2020.

\bibitem[Ghorbani et~al.(2019)Ghorbani, Krishnan, and Xiao]{Ghorbani2019AnII}
Ghorbani, B., Krishnan, S., and Xiao, Y.
\newblock An investigation into neural net optimization via hessian eigenvalue
  density.
\newblock In \emph{ICML}, 2019.

\bibitem[He et~al.(2016)He, Zhang, Ren, and Sun]{He2016DeepRLResnet}
He, K., Zhang, X., Ren, S., and Sun, J.
\newblock Deep residual learning for image recognition.
\newblock \emph{2016 IEEE Conference on Computer Vision and Pattern Recognition
  (CVPR)}, pp.\  770--778, 2016.

\bibitem[Ho et~al.(2019)Ho, Kalchbrenner, Weissenborn, and
  Salimans]{Ho2019AxialAITransformer1D}
Ho, J., Kalchbrenner, N., Weissenborn, D., and Salimans, T.
\newblock Axial attention in multidimensional transformers.
\newblock \emph{ArXiv}, abs/1912.12180, 2019.

\bibitem[Ioffe(2017)]{Ioffe2017BatchRT}
Ioffe, S.
\newblock Batch renormalization: Towards reducing minibatch dependence in
  batch-normalized models.
\newblock \emph{ArXiv}, abs/1702.03275, 2017.

\bibitem[Ioffe \& Szegedy(2015)Ioffe and Szegedy]{BatchNorm}
Ioffe, S. and Szegedy, C.
\newblock Batch normalization: Accelerating deep network training by reducing
  internal covariate shift.
\newblock \emph{ArXiv}, abs/1502.03167, 2015.

\bibitem[Jastrzebski et~al.(2020)Jastrzebski, Szymczak, Fort, Arpit, Tabor,
  Cho, and Geras]{Jastrzebski2020TheBP}
Jastrzebski, S., Szymczak, M., Fort, S., Arpit, D., Tabor, J., Cho, K., and
  Geras, K.~J.
\newblock The break-even point on optimization trajectories of deep neural
  networks.
\newblock \emph{ArXiv}, abs/2002.09572, 2020.

\bibitem[Kessy et~al.(2018)Kessy, Lewin, and Strimmer]{Kessy2018OptimalWA}
Kessy, A. M.~N., Lewin, A., and Strimmer, K.
\newblock Optimal whitening and decorrelation.
\newblock \emph{The American Statistician}, 72:\penalty0 309 -- 314, 2018.

\bibitem[Kohler et~al.(2018)Kohler, Daneshmand, Lucchi, Zhou, Neymeyr, and
  Hofmann]{Kohler2018TowardsAT}
Kohler, J.~M., Daneshmand, H., Lucchi, A., Zhou, M., Neymeyr, K., and Hofmann,
  T.
\newblock Towards a theoretical understanding of batch normalization.
\newblock \emph{ArXiv}, abs/1805.10694, 2018.

\bibitem[Krizhevsky(2009{\natexlab{a}})]{Krizhevsky2009LearningML}
Krizhevsky, A.
\newblock Learning multiple layers of features from tiny images.
\newblock 2009{\natexlab{a}}.

\bibitem[Krizhevsky(2009{\natexlab{b}})]{cifar10}
Krizhevsky, A.
\newblock Learning multiple layers of features from tiny images.
\newblock 2009{\natexlab{b}}.

\bibitem[Laurent et~al.(2017)Laurent, Ballas, and
  Vincent]{Laurent2017RecurrentNP}
Laurent, C., Ballas, N., and Vincent, P.
\newblock Recurrent normalization propagation.
\newblock In \emph{ICLR}, 2017.

\bibitem[LeCun et~al.(1992)LeCun, Simard, and
  Pearlmutter]{LeCun1992AutomaticLR}
LeCun, Y., Simard, P.~Y., and Pearlmutter, B.~A.
\newblock Automatic learning rate maximization by on-line estimation of the
  hessian's eigenvectors.
\newblock In \emph{NIPS 1992}, 1992.

\bibitem[Lewkowycz et~al.(2020)Lewkowycz, Bahri, Dyer, Sohl-Dickstein, and
  Gur-Ari]{lewkowycz2020large}
Lewkowycz, A., Bahri, Y., Dyer, E., Sohl-Dickstein, J., and Gur-Ari, G.
\newblock The large learning rate phase of deep learning: the catapult
  mechanism, 2020.

\bibitem[Luo et~al.(2019)Luo, Wang, Shao, and Peng]{Luo2019TowardsUR}
Luo, P., Wang, X., Shao, W., and Peng, Z.
\newblock Towards understanding regularization in batch normalization.
\newblock \emph{ArXiv}, abs/1809.00846, 2019.

\bibitem[Poole et~al.(2016)Poole, Lahiri, Raghu, Sohl-Dickstein, and
  Ganguli]{Poole2016ExponentialEI}
Poole, B., Lahiri, S., Raghu, M., Sohl-Dickstein, J., and Ganguli, S.
\newblock Exponential expressivity in deep neural networks through transient
  chaos.
\newblock In \emph{NIPS}, 2016.

\bibitem[Salimans \& Kingma(2016)Salimans and Kingma]{Salimans2016WeightNA}
Salimans, T. and Kingma, D.~P.
\newblock Weight normalization: A simple reparameterization to accelerate
  training of deep neural networks.
\newblock \emph{ArXiv}, abs/1602.07868, 2016.

\bibitem[Sankararaman et~al.(2019)Sankararaman, De, Xu, Huang, and
  Goldstein]{Sankararaman2019GradConfusion}
Sankararaman, K.~A., De, S., Xu, Z., Huang, W.~R., and Goldstein, T.
\newblock The impact of neural network overparameterization on gradient
  confusion and stochastic gradient descent.
\newblock \emph{ArXiv}, abs/1904.06963, 2019.

\bibitem[Santurkar et~al.(2018)Santurkar, Tsipras, Ilyas, and
  Madry]{Santurkar2018HowDB}
Santurkar, S., Tsipras, D., Ilyas, A., and Madry, A.
\newblock How does batch normalization help optimization?
\newblock In \emph{NeurIPS}, 2018.

\bibitem[Schoenholz et~al.(2016)Schoenholz, Gilmer, Ganguli, and
  Sohl-Dickstein]{schoenholz2016deep}
Schoenholz, S.~S., Gilmer, J., Ganguli, S., and Sohl-Dickstein, J.
\newblock Deep information propagation.
\newblock \emph{arXiv preprint arXiv:1611.01232}, 2016.

\bibitem[Shen et~al.(2020)Shen, Yao, Gholami, Mahoney, and
  Keutzer]{shen2020rethinking}
Shen, S., Yao, Z., Gholami, A., Mahoney, M., and Keutzer, K.
\newblock Rethinking batch normalization in transformers, 2020.

\bibitem[Singh \& Krishnan(2019)Singh and Krishnan]{Singh2019FilterRN}
Singh, S. and Krishnan, S.
\newblock Filter response normalization layer: Eliminating batch dependence in
  the training of deep neural networks.
\newblock \emph{ArXiv}, abs/1911.09737, 2019.

\bibitem[Ulyanov et~al.(2016)Ulyanov, Vedaldi, and Lempitsky]{InstanceNorm}
Ulyanov, D., Vedaldi, A., and Lempitsky, V.~S.
\newblock Instance normalization: The missing ingredient for fast stylization.
\newblock \emph{ArXiv}, abs/1607.08022, 2016.

\bibitem[Vaswani et~al.(2017)Vaswani, Shazeer, Parmar, Uszkoreit, Jones, Gomez,
  Kaiser, and Polosukhin]{Vaswani2017AttentionIA}
Vaswani, A., Shazeer, N., Parmar, N., Uszkoreit, J., Jones, L., Gomez, A.~N.,
  Kaiser, L., and Polosukhin, I.
\newblock Attention is all you need.
\newblock \emph{ArXiv}, abs/1706.03762, 2017.

\bibitem[Wu \& He(2018)Wu and He]{GroupNorm}
Wu, Y. and He, K.
\newblock Group normalization.
\newblock In \emph{ECCV}, 2018.

\bibitem[Xiao et~al.(2018)Xiao, Bahri, Sohl-Dickstein, Schoenholz, and
  Pennington]{Xiao2018DynamicalIA}
Xiao, L., Bahri, Y., Sohl-Dickstein, J., Schoenholz, S.~S., and Pennington, J.
\newblock Dynamical isometry and a mean field theory of cnns: How to train 10,
  000-layer vanilla convolutional neural networks.
\newblock \emph{ArXiv}, abs/1806.05393, 2018.

\bibitem[Xiao et~al.(2019)Xiao, Pennington, and
  Schoenholz]{Xiao2019DisentanglingTA}
Xiao, L., Pennington, J., and Schoenholz, S.~S.
\newblock Disentangling trainability and generalization in deep learning.
\newblock \emph{ArXiv}, abs/1912.13053, 2019.

\bibitem[Xiong et~al.(2020)Xiong, Yang, He, Zheng, xin Zheng, Xing, Zhang, Lan,
  Wang, and Liu]{Xiong2020OnLN}
Xiong, R., Yang, Y., He, D., Zheng, K., xin Zheng, S., Xing, C., Zhang, H.,
  Lan, Y., Wang, L.-W., and Liu, T.-Y.
\newblock On layer normalization in the transformer architecture.
\newblock \emph{ArXiv}, abs/2002.04745, 2020.

\bibitem[Yang et~al.(2019)Yang, Pennington, Rao, Sohl-Dickstein, and
  Schoenholz]{Yang2019AMF}
Yang, G., Pennington, J., Rao, V., Sohl-Dickstein, J., and Schoenholz, S.~S.
\newblock A mean field theory of batch normalization.
\newblock \emph{ArXiv}, abs/1902.08129, 2019.

\bibitem[Zagoruyko \& Komodakis(2016)Zagoruyko and
  Komodakis]{Zagoruyko2016WideRN}
Zagoruyko, S. and Komodakis, N.
\newblock Wide residual networks.
\newblock \emph{ArXiv}, abs/1605.07146, 2016.

\end{thebibliography}

\clearpage
\newpage
\medskip

\appendix

\setcounter{figure}{0}
\setcounter{table}{0}

\renewcommand{\thefigure}{App.\arabic{figure}}
\renewcommand{\thetable}{App.\arabic{table}}

\section{Failed experiments} \label{sec:failed_exps}

In addition to the analysis and successful techniques we introduced above, we tried several other seemingly sensible approaches to match the statistical properties of Batch Norm. The following approaches failed to improve performance.

\begin{figure}[h]
\centering
    \includegraphics[width=\linewidth]{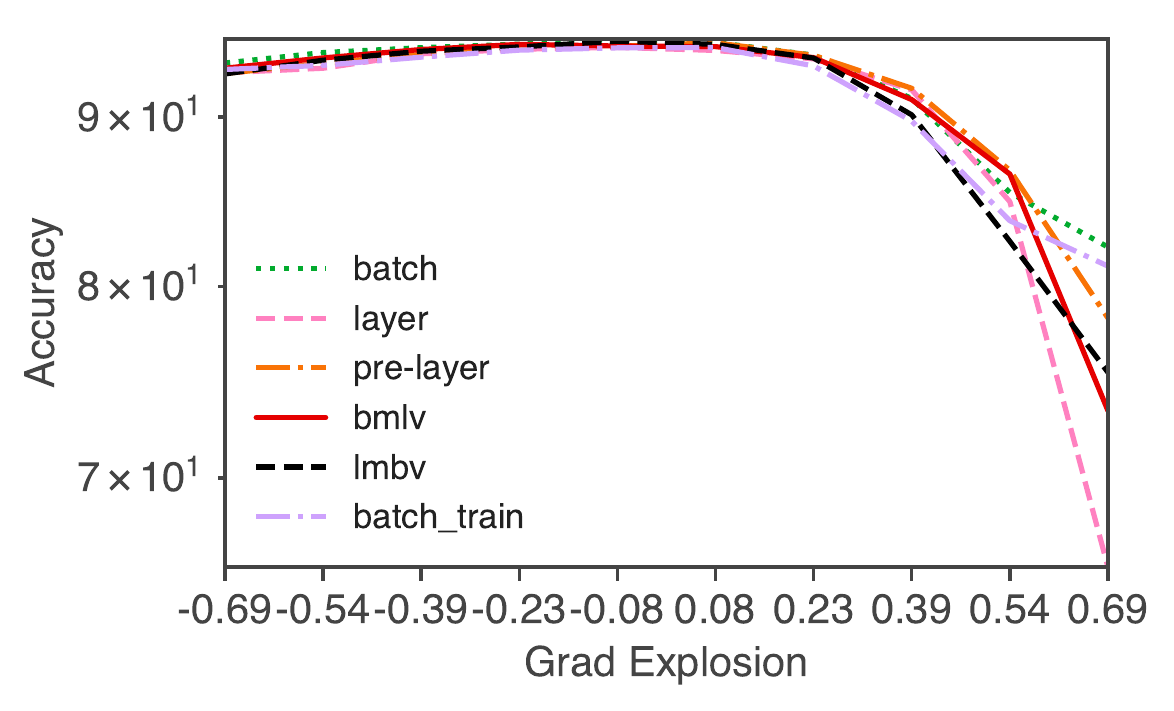}
 \caption{Induced gradient explosion. This mimics Batch Norm's gradient explosion by introducing a scalar multiplier to the gradients at each layer. Note that the optimal gradient explosion/vanishing just turns out to be $\sim$1 for all the techniques with performance rapidly decreasing when the gradient explosion is high.}
    \label{fig:grad_explosion}
\end{figure}

\begin{figure}[h]
\centering
    \includegraphics[width=\linewidth]{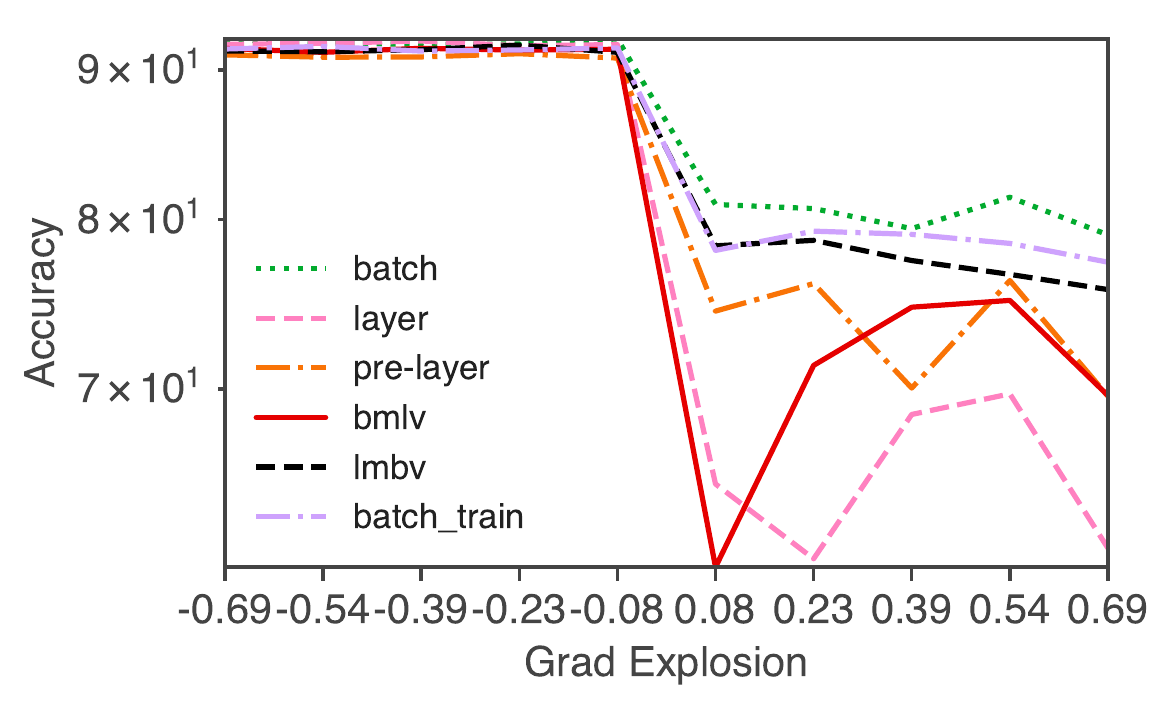}
 \caption{Induced gradient explosion with a schedule. This mimics Batch Norm's gradient explosion by introducing a scalar multiplier to the gradients at each layer, only for the first 50 steps of training. We arrive at this number by looking at the change in gradient norms of networks with Batch Norm through training. High gradient explosion hurts performance for all the techniques, while with this schedule, networks with slightly vanishing gradients perform consistently with the optimal ($\sim$1)}
    \label{fig:grad_explosion_schedule}
\end{figure}

\paragraph{Artificial gradient explosion} We have seen that Batch Norm causes more gradient explosion with ReLU networks than other techniques and this was also studied by \citet{Yang2019AMF}. To simulate this, we forcefully insert a scaling factor at each layer to increase the scale of the gradients through the network (Figure \ref{fig:grad_explosion}). Through training, there is a phase where gradient explosion quickly reduces in the early stages. We mimicked this by linearly decreasing the scaling factor until 50 steps and then turning it off (Figure \ref{fig:grad_explosion_schedule}). However, both these techniques failed to improve performance. In fact, we note that the best performing scaling was the default.

\paragraph{Simulating batch noise} Batch Normalization seems to rely on the regularization effects introduced by the differences in training and test-time behavior for better generalization. These differences are mainly due to the noise in estimating the mean and variance for every batch while training. We tried to mimic this by adding noise to the inputs after normalization -- i.e setting pre-activations to $\tilde{z}^l + \epsilon$ where $\epsilon \in \mathcal{N}(0,\sigma)$. Further, we used the same random seed through the network while sampling $\epsilon$, to capture the dependencies between minibatch noise at different layers. This failed to improve generalization during test-time.

\paragraph{Artificial chaos} From the observed informational propagation properties of Batch Norm, one could imagine designing architectures that induce chaos at initialization the same way Batch Norm does, and we tried these methods:
\begin{itemize}
    \item Signed ReLU activation:\\ $\phi(x) = \alpha * \dfrac{1+sgn(x)}{2} + max(0,x)$
    \item Normalized Affine Operation:\\ $x^l = \phi\left(\dfrac{Wx}{\sqrt{W^2x^2}}\right)$
    \item TLU (Thresholded Linear Unit) proposed by \citet{Singh2019FilterRN}
\end{itemize}

\paragraph{Targeting Hessian conditioning} To condition the Hessian very similarly to Batch Norm without using it every layer, we used batch mean subtraction and/or batch variance division operations only at the final logits layer\footnote{Hinted at by https://myrtle.ai/learn/how-to-train-your-resnet-7-batch-norm/}. Although this conditions the Hessians similarly, we noticed a drop in generalization.\\
\paragraph{Regularized learning of variance} One of our proposed normalization schemes, RegNorm, also lends itself to a natural extension where with another regularizer ($r(y) = \mathbb{E}_{a,b}\left[ \sum_i (z^a_iz^b_i - 1) \right]$), one might push the variance of the batch to be 1 and totally avoid the cost of normalization during inference. This caused training to be unstable and did not improve generalization.

\section{Regularizer is minimized at batch-mean 0} \label{app:regularizer_proof}
The regularizer we propose (Section \ref{sec:reg_norm}) is of the form: 
\begin{align*}
    r(\Bar{z}) &= \mathbb{E}_{a,b}\left[ (\sum_i (\Bar{z}^a_i + \Bar{z}^b_i)^2 - 2) \right]
\end{align*}
where $\Bar{z} = \dfrac{z^l}{\sigma^i(\cdot)}$, $a, b \in Batch, i \in (H, W, C)$. Let $N_l$ be the number of neurons at layer $l$ (W*H*C).
\begin{align*}
    r(\Bar{z}) &= \mathbb{E}_{a,b}\left[ \sum_i ((\Bar{z}^a_i)^2 + (\Bar{z}^b_i)^2 + 2\Bar{z}^a_i\Bar{z}^b_i - 2) \right]\\
    &= 2\mathbb{E}_a \left[ \sum_i((\Bar{z}^a_i)^2) \right] + 2\sum_i\mathbb{E}_{a,b}[\Bar{z}^a_i\Bar{z}^b_i] - 2N_l
\end{align*}
Note that $\Bar{z^a_i} = \dfrac{z^a_i}{\sqrt{(1/N_l)\sum_j(z^a_j)^2}}$, and $\mathbb{E}_{a,b}[\Bar{z}^a_i\Bar{z}^b_i] = \mathbb{E}_a[(\Bar{z}^a_i)^2]$
\begin{align*}
    r(\Bar{z}) &= 2\mathbb{E}_i\left[ \sum_i\left( \dfrac{(z^a_i)^2}{(1/N_l)\sum_j(z^a_j)^2)} \right) \right] + 2\sum_i\mathbb{E}[(\Bar{z}^a_i)^2]\\& \qquad - 2N_l\\
    &= 2N_l + 2\sum_i\mathbb{E}_a[(\Bar{z}^a_i)^2] - 2N_l\\
    &= 2\sum_i\mathbb{E}_a[(\Bar{z}^a_i)^2]
\end{align*}
Setting $r(\Bar{z})$ to 0, we see that this is only achieved when $\mathbb{E}_a[(\Bar{z}^a_i)^2]$ is 0 $\forall a \in Batch$ i.e, when the means of all the inputs the batch are 0, as required.
\small

\section{Further properties and ablation studies} \label{appendix:ablation}
\begin{figure}[h]
    \centering
    \includegraphics[width=\linewidth]{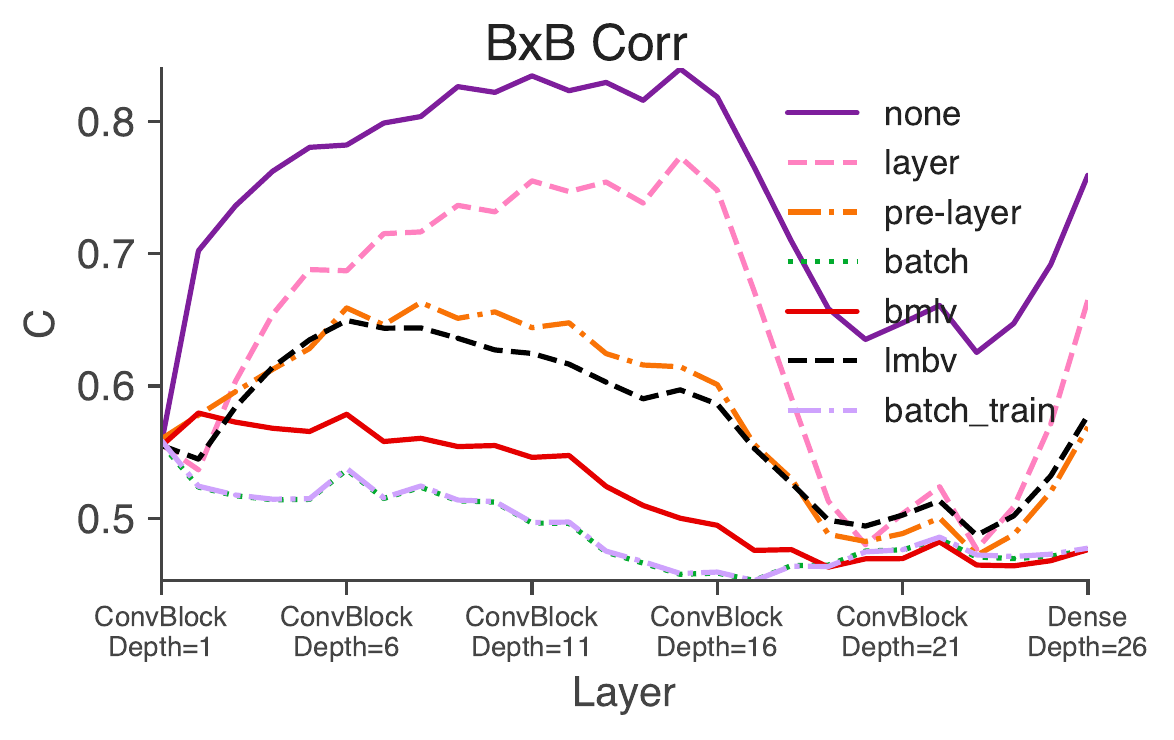}
 \caption{Correlation of inputs between batches through a 26-layer WideResnet with skip connections. Batch Norm cases the input representations to be more decorrelated through depth than other techniques. However, with skip-connections, inputs are never completely decorrelated.}
    \label{fig:wrn_bxb}
\end{figure}
\begin{figure}[h]
\centering
    \includegraphics[width=0.5\textwidth]{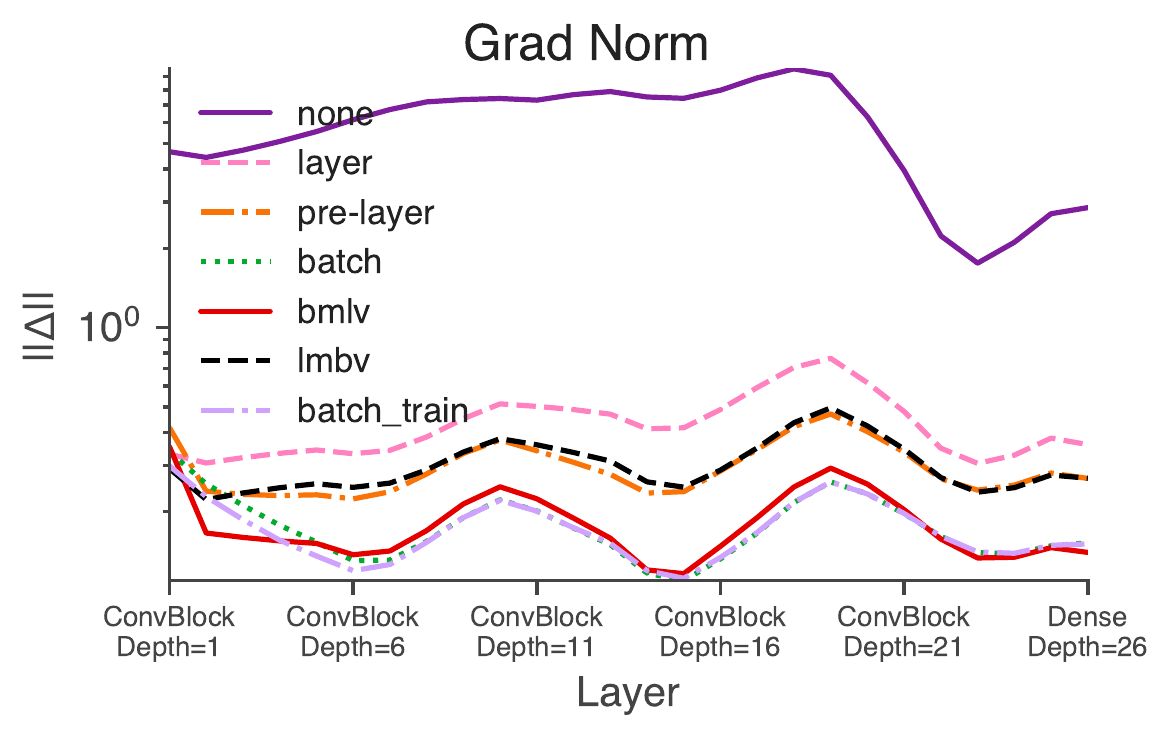}
 \caption{L2-Norm of the gradients of the parameters of all the layers in a WideResnet without skip connections. With skip connections, all the normalization techniques avoid gradient explosion with No Norm causing the most explosion.}
    \label{fig:grad_norm}
\end{figure}

\begin{figure}[h]
\centering
    \includegraphics[width=\linewidth]{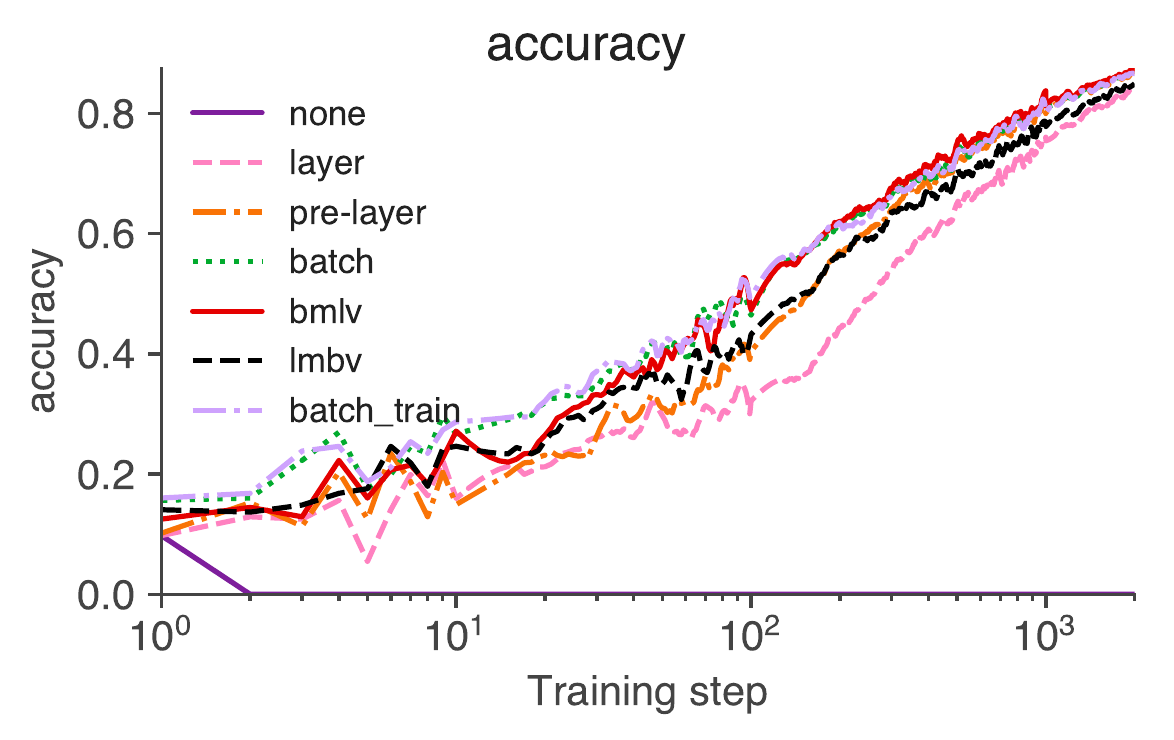}
 \caption{Accuracy of a 26-layer WideResnet ReLU network with skip connections through the first 2000 steps of training on Cifar10. Note that No Norm fails to train at this depth with high learning rates, and most normalization schemes are able to reach 50\% accuracy by 500 steps into training.}
    \label{fig:train_accuracy}
\end{figure}

\begin{figure}[h]
    \centering
    \includegraphics[width=\linewidth]{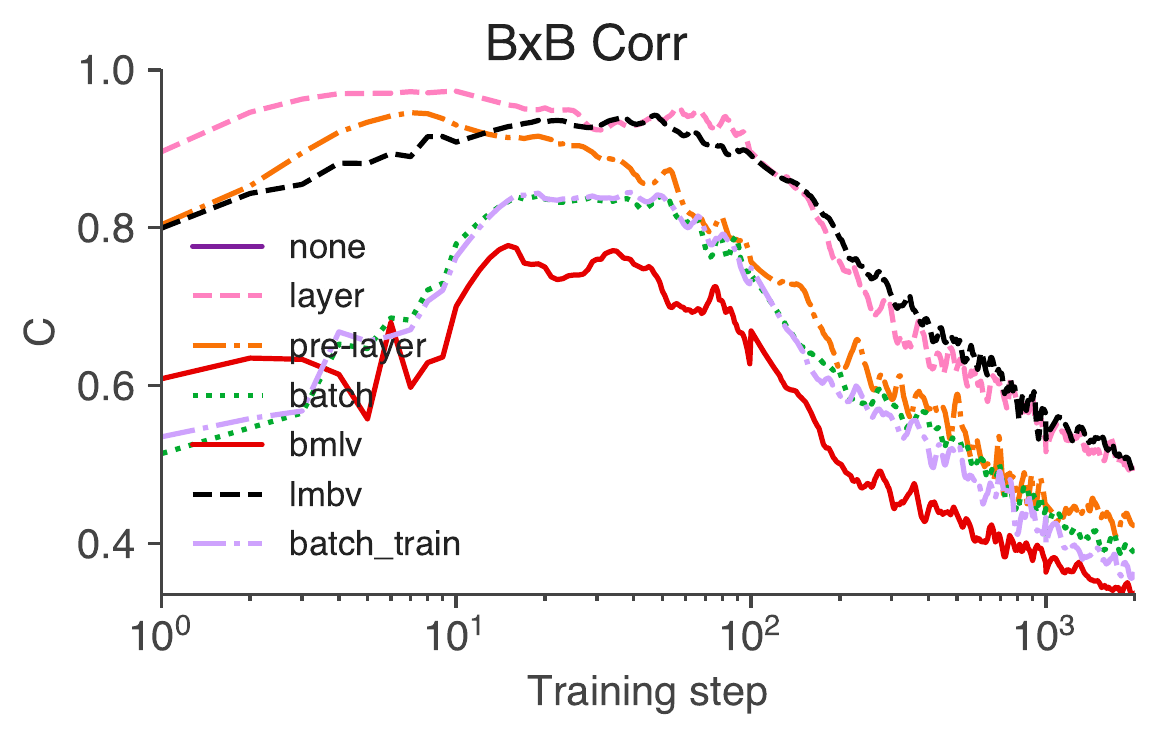}
    \caption{Correlation between output representations of similar minibatches at the last pre-activation layer through a 26-layer WideResnet ReLU network with skip-connections through the first 2000 steps of training on Cifar10. Even with skip connections, these networks undergo phase shifts where the representations get more correlated early in training and later on are more decorrelated. Note that Batch Norm (and BMLV) are the most decorrelated while PreLayerNorm (and LMBV) is more decorrelated than Layer Norm.}
    \label{fig:bxb_skip}
\end{figure}

\begin{figure*}[h]
    \centering
    \includegraphics[width=\textwidth]{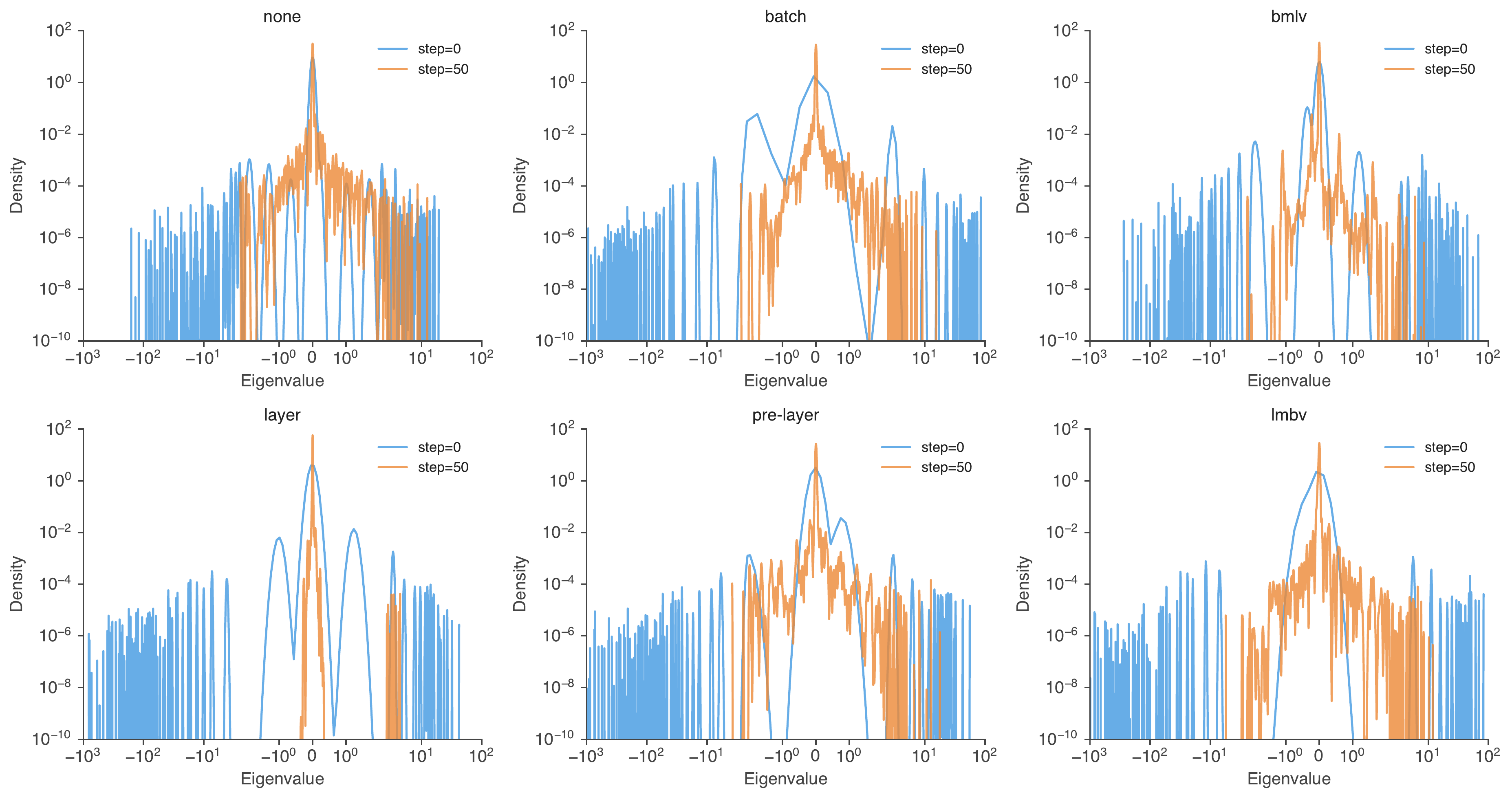}
    \caption{Density of Eigenvalues of the Hessian (cross-entropy) at initialization and 50 steps through training (we used the same learning rate for all techniques) for a 26-layer WideResnet with ReLU activations and no skip connections. With skip connections, we notice that all the normalization schemes seem to have similar spectra.}
    \label{fig:density_skip}
\end{figure*}

\begin{figure}[h]
\centering
    \includegraphics[width=\linewidth]{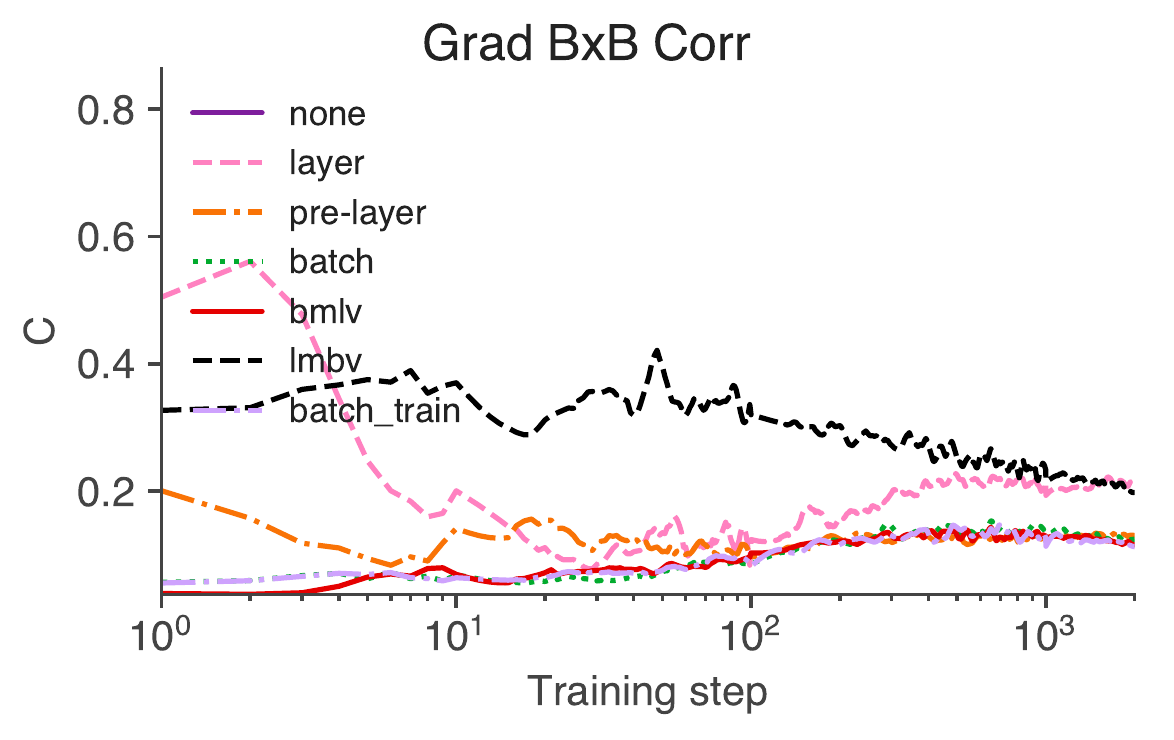}
 \caption{Correlation between gradients of similar minibatches w.r.t parameters at the last pre-activation layer through a 26-layer WideResnet ReLU network with skip-connections through the first 2000 steps of training on Cifar10. With skip connections, most techniques are able to avoid gradient confusion early in training, except LMBV which does so to a lesser degree.}
    \label{fig:grad_bxb_skip}
\end{figure}

\begin{figure}[h]
\centering
    \includegraphics[width=\linewidth]{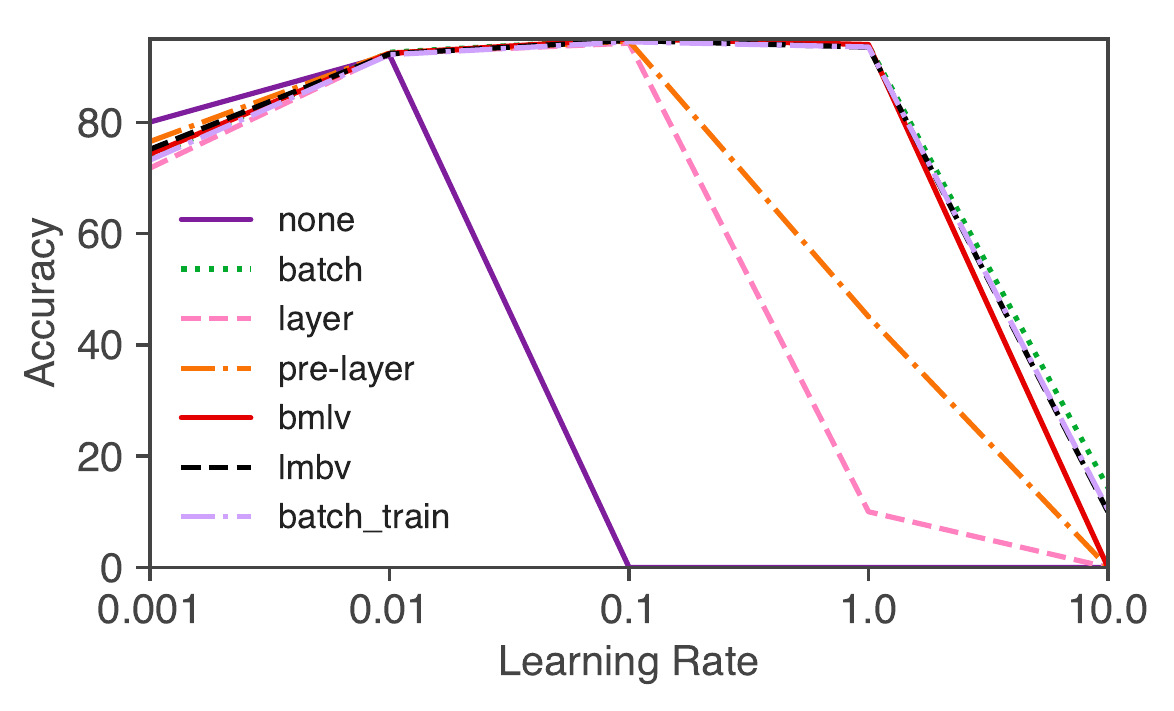}
 \caption{Effect of using a wide range of learning rates under different normalization techniques. Batch Norm and BMLV allow using larger stable learning rates while PreLayerNorm allows higher learning rates than Layer Norm.}
    \label{fig:lr}
\end{figure}
\begin{figure}[h]
\centering
    \includegraphics[width=\linewidth]{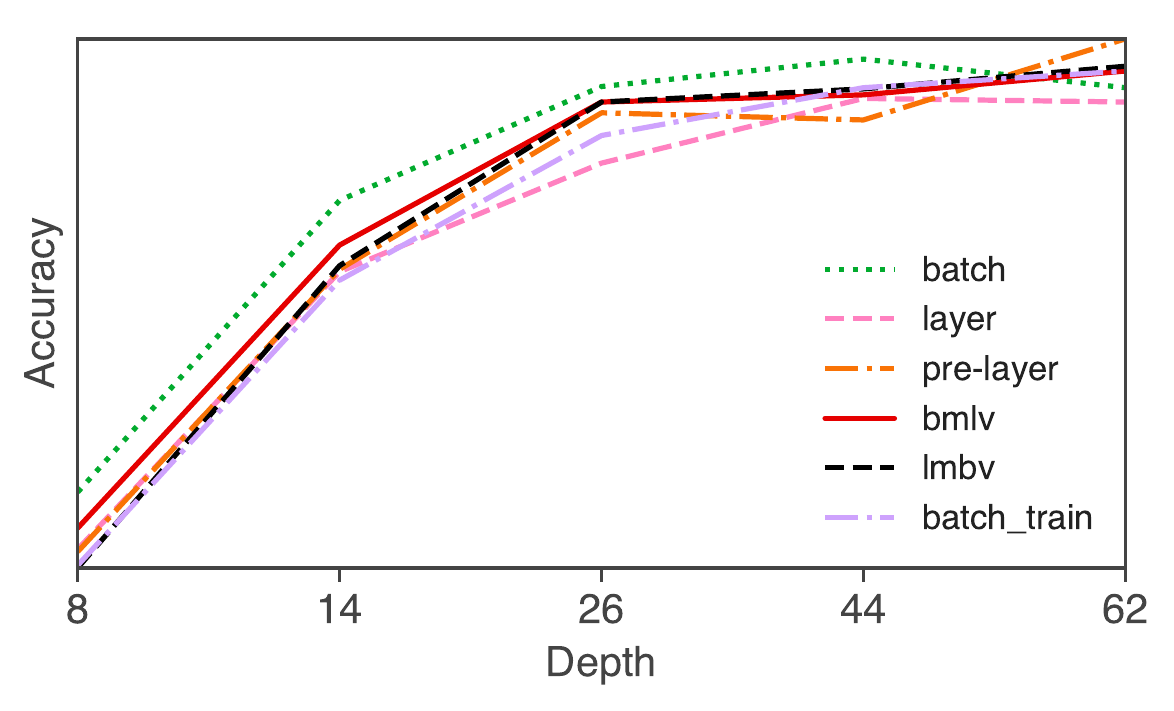}
 \caption{Effect of increasing depth under different normalization techniques. Most of the normalization techniques used allow training deeper networks, especially with skip connections. Without skip connections, using normalization is imperative for training deeper networks.}
    \label{fig:depth}
\end{figure}

\begin{figure}[h]
\centering
    \includegraphics[width=\linewidth]{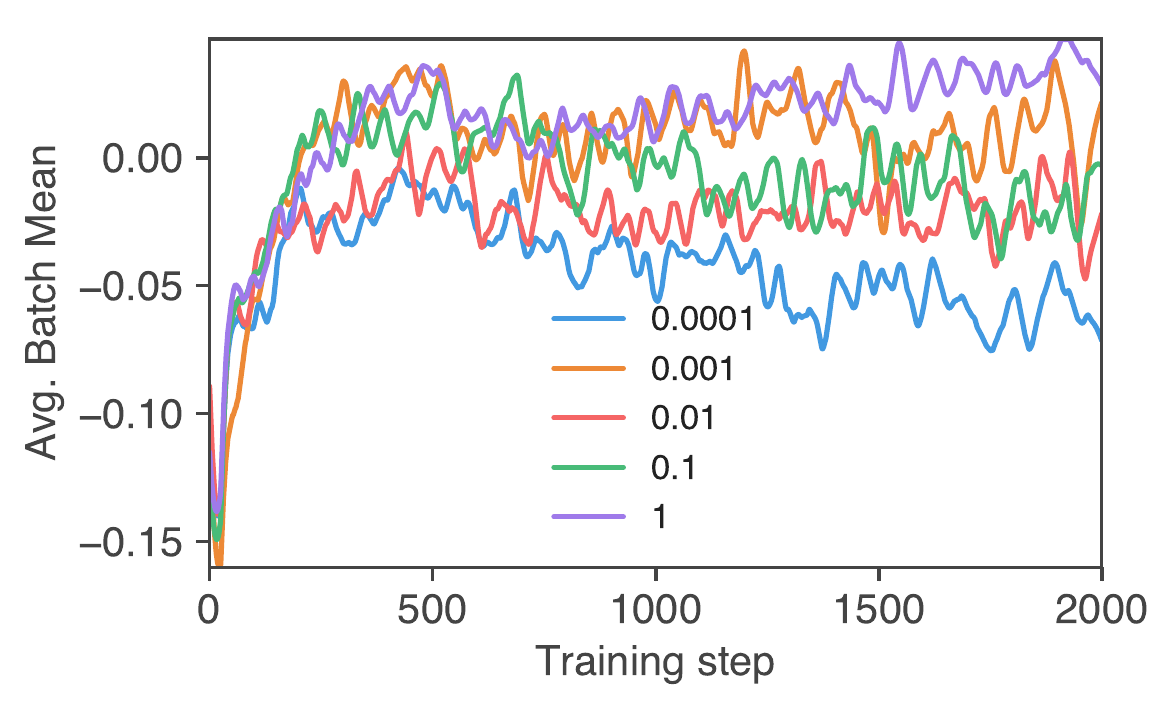}
 \caption{Effect of using RegNorm on the batch mean of the pre-activations at a random layer in a 26-layer WideResnet. This figure plots the average (across all the channels) batch mean for the first 2000 steps through training against different values of the regularization co-efficient. We observed that larger co-efficients push the mean across the batch to be closer to 0. However, if the co-efficient was too large, it hurt the trainability of the network.}
    \label{fig:reg_coeff}
\end{figure}

\end{document}